\documentclass[journal]{IEEEtran}
\usepackage{cite}
\usepackage{amsmath,amssymb,amsfonts}
\usepackage{graphicx}
\usepackage{textcomp}
\usepackage{xcolor}
\usepackage{verbatim}

\usepackage{multirow}
\usepackage{subcaption}
\usepackage{booktabs}
\usepackage[algo2e,linesnumbered,lined,boxed,commentsnumbered,ruled]{algorithm2e}
\usepackage{array}
\usepackage[caption=false,font=normalsize,labelfont=sf,textfont=sf]{subfig}
\usepackage{stfloats}
\usepackage{url}

\def\BibTeX{{\rm B\kern-.05em{\sc i\kern-.025em b}\kern-.08em
    T\kern-.1667em\lower.7ex\hbox{E}\kern-.125emX}}
\begin{document}

\title{Graph2Region: Efficient Graph Similarity Learning with Structure and Scale Restoration}
\markboth{Journal of \LaTeX\ Class Files,~Vol.~14, No.~8, August~2021}
{Shell \MakeLowercase{\textit{et al.}}: A Sample Article Using IEEEtran.cls for IEEE Journals}
\author{Zhouyang Liu,
        Yixin Chen$^\dag$,
        Ning Liu$^\dag$,
        Jiezhong He,
        Dongsheng Li
\thanks{Zhouyang Liu, Yixin Chen, Jiezhong He, and Dongsheng Li are with the College of Computer Science and Technology, National University of Defense Technology, Changsha, Hunan, China (e-mail: \{liuzhouyang20, chenyixin, jiezhonghe, dsli\}@nudt.edu.cn). Ning Liu is with Information Support Force Engineering University, Wuhan, Hubei, China (e-mail: liuning17a@nudt.edu.cn).}
\thanks{This work is supported in part by National Key Research and Development Program of China (No.
2023YFB4502300), the National Natural Science Foundation of China under grants (Nos. 62402503,
62025208 and 62421002) and State Administration of Science Technology and Industry for National Defense Foundation
under Grant WDZC20255290104. \emph{(Corresponding authors: Yixin Chen, Ning Liu).}}}

\maketitle

\begin{abstract}
Graph similarity is critical in graph-related tasks such as graph retrieval, where metrics like maximum common subgraph (MCS) and graph edit distance (GED) are commonly used. However, exact computations of these metrics are known to be NP-Hard. Recent neural network-based approaches approximate the similarity score in embedding spaces to alleviate the computational burden, but they either involve expensive pairwise node comparisons or fail to effectively utilize structural and scale information of graphs. To tackle these issues, we propose a novel geometric-based graph embedding method called \textsc{Graph2Region} (\textsc{G2R}). \textsc{G2R} represents nodes as closed regions and recovers their adjacency patterns within graphs in the embedding space. By incorporating the node features and adjacency patterns of graphs, \textsc{G2R} summarizes graph regions, i.e., graph embeddings, where the shape captures the underlying graph structures and the volume reflects the graph size. Consequently, the overlap between graph regions can serve as an approximation of MCS, signifying similar node regions and adjacency patterns. We further analyze the relationship between MCS and GED and propose using disjoint parts as a proxy for GED similarity. This analysis enables concurrent computation of MCS and GED, incorporating local and global structural information. Experimental evaluation highlights \textsc{G2R}'s competitive performance in graph similarity computation. It achieves up to a 60.0\% relative accuracy improvement over state-of-the-art methods in MCS similarity learning, while maintaining efficiency in both training and inference. Moreover, \textsc{G2R} showcases remarkable capability in predicting both MCS and GED similarities simultaneously, providing a holistic assessment of graph similarity. Code available at \url{https://github.com/liuzhouyang/Graph2Region}.
\end{abstract}

\begin{IEEEkeywords}
Graph Similarity Computation, Graph Representation Learning
\end{IEEEkeywords}

\section{Introduction}
\IEEEPARstart{G}{raphs} are essential for modeling interactions between entities. Graph similarity metrics such as Maximum Common Subgraph (MCS) and Graph Edit Distance (GED), which quantify the structural and characteristic resemblance between graph pairs, provide valuable insights into tasks like graph similarity search \cite{database, database2, database3, database4}, network analysis \cite{na, na2} and drug discovery \cite{drug, drug2}. However, the exact computation of these metrics is time-consuming due to their NP-Hard nature \cite{BUNKE1997689}. Even state-of-the-art classic algorithms struggle to efficiently compute GED for graphs with more than sixteen nodes \cite{BLUMENTHAL202046}, limiting the practical utility. To reduce computational complexity and response time, recent studies have explored neural networks to directly predict similarity scores, enabling more efficient handling of a large number of small-sized queries in graph retrieval task.

In MCS and GED computations, graph similarity is primarily determined by structural and feature information induced by nodes. As a result, fine-grained pairwise comparison between node embeddings has become a prevailing neural-based strategy \cite{Bai2019SimGNNAN,Li2019GraphMN,graphsim, Doan2021InterpretableGS, Roy2022MaximumCS, Roy2022InterpretableNS,deepsim}. These methods compare nodes from both graphs to estimate similarity. Despite the considerable acceleration compared with classic algorithms, they remain time-consuming during both training and inference stages due to the tedious comparison. Even worse, the irregular nature of graphs necessitates padding compared pairs to a uniform size, which not only increases the memory overhead on GPUs but also incurs additional computations.

\begin{figure}[t]
    \centering
    \includegraphics[width=\linewidth]{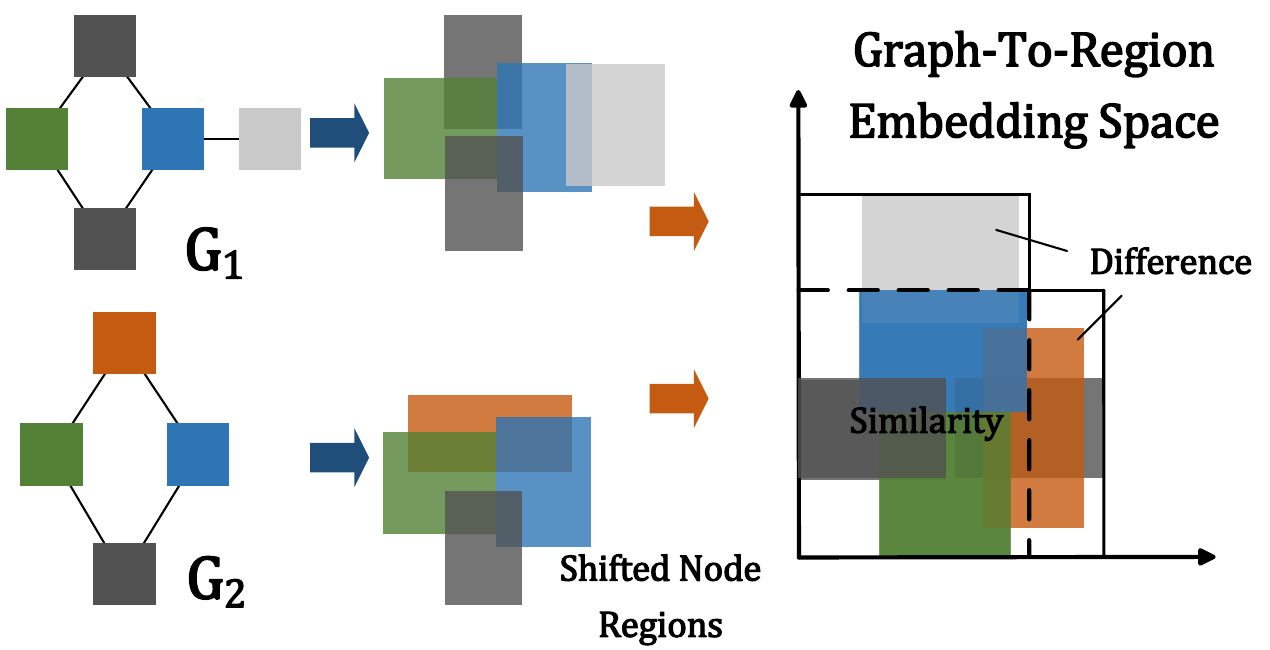}
    \caption{Node regions are shifted to reflect the adjacency pattern within graphs. The graph regions of $G_1$ and $G_2$ have a substantial overlap, which signifies that they exhibit comparable node regions and similar connectivity patterns.}
    \label{fig:intuition}
\end{figure}

To overcome computational bottlenecks in pairwise node comparison, researchers propose using graph-level embeddings for coarse-grained estimation \cite{Li2019GraphMN,zhuo2022efficient,ranjan2022greed}. These approaches represent graphs as single points in the embedding space, thereby reducing the number of elements to be compared and alleviating the computational burden. However, this oversimplification introduces a potential trade-off as it overlooks the explicit utilization of structural and graph scale information. Neglecting these factors can compromise the expressive power of graph embeddings and risk treating diverse graphs as identical. 

To address the above limitations, we propose a novel graph embedding model that explicitly transfers \underline{G}raph structure and scale information \underline{to} geometric \underline{R}egions for efficient graph similarity computation, namely \textsc{Graph2Region} (\textsc{G2R} for short). \textsc{G2R} introduces an innovative \emph{Multi-sink Propagation} mechanism, which computes flow paths from nodes to multiple target nodes called sinks. This mechanism captures the similarity in paths among nearby nodes, generating relative positions that reflect node proximity, thereby capturing crucial structural information. Furthermore, inspired by box embeddings \cite{vilnis2018probabilistic, Li2018SmoothingTG, dasgupta2020improving}, where hyperrectangles are used to represent concepts with volume as a measure of probability. \textsc{G2R} formulate node representations as closed regions in the embedding space toward the common ordinate origin. It shifts these regions to their relative positions to align with the adjacency patterns, thereby restoring the graph structures. \textsc{G2R} then generates graph regions by leveraging the shifted node regions, and constraining the volume of the graph regions to reflect the scale of graphs. In this way, \textsc{G2R} consolidates both structural and scale information, resulting in a more expressive representation.

Under such modeling, \textsc{G2R} explicitly encode the node features and adjacency patterns of graphs while implicitly aligning them within the embedding space. This approach effectively captures the essential similarities between graphs, eliminating the need for explicit fine-grained comparisons, enabling efficient and accurate graph similarity computation. Specifically, \textsc{G2R} treats the overlap between two graph regions as an approximation of the maximum shared substructure (MCS), and regards disjoint parts as their difference, as illustrated in Fig. \ref{fig:intuition}. \textsc{G2R} predicts the MCS similarity based on the shape and volume of the overlapped region. Furthermore, inspired by prior theoretical work on graph matching \cite{BUNKE1997689}, we analyze the relationship between MCS and GED, and leverage the disjoint regions as a proxy for GED similarity. This decoupling of input for prediction allows \textsc{G2R} to simultaneously estimate MCS and GED similarities, resulting in a more comprehensive assessment of graph similarity. 

We summarize our main contributions as follows:
\begin{itemize}
    \item We propose a geometric-based graph embedding model for graph similarity learning, which explicitly restores the structural and scale information of the original graphs, enhancing the expressiveness of the graph embeddings.
    \item We introduce a \emph{Multi-sink Propagation} mechanism, which transforms the relative positional encoding problem into sequence similarity learning, capturing node proximity as crucial structural information.
    \item We explore the possibility of computing MCS and GED concurrently and validate this possibility theoretically and empirically. This approach ensures a more holistc assessment of graph similarity by integrating both local and global structural information. 
\end{itemize}

Empirical results on thirteen datasets validate the effectiveness of \textsc{G2R} on MCS similarity learning,  showing a relative accuracy improvement ranging from 7.7\% to 60.0\%. The results also demonstrate remarkable transferability while maintaining time efficiency. Moreover, extensive results on three popular GED datasets provide a comprehensive analysis of \textsc{G2R}'s key components, highlighting \textsc{G2R}'s unique ability to predict MCS and GED similarities simultaneously.

\section{Related Work} \label{sec:related_work}
Our research focuses on deep graph similarity learning and draws inspiration from geometric representation learning. In the following, we briefly review representative works within these domains.

\textbf{Deep Graph Similarity Learning.} Deep graph similarity learning refers to using neural networks to learn the similarity between graphs in embedding spaces. Pioneering works in this domain include SimGNN \cite{Bai2019SimGNNAN} and GMN \cite{Li2019GraphMN}, which introduced cross-graph node comparison techniques. These methods demonstrated superior performance compared with classic GED algorithms and graph-level embedding approaches, such as GEN \cite{Li2019GraphMN}. Since then, there has been a surge in fine-grained approaches \cite{graphsim,Doan2021InterpretableGS, Roy2022MaximumCS,vldb,h2mn,deepsim}. These approaches are lightweight in embedding generation but rely heavily on fine-grained node comparison, resulting in time-consuming training and inference phases. The irregular nature of graphs also increases the memory overhead on GPUs. Recent studies have focused on coarse-grained estimation methods to compute graph similarities efficiently. These methods rely exclusively on graph-level embeddings during the inference phase \cite{zhuo2022efficient, ranjan2022greed, qin2021slow}. However, they may sacrifice important structural and scale information that characterizes graphs.

Despite the importance of scale and structural information, existing methods commonly fall short of effectively incorporating it. SimGNN uses histogram features to reflect graph scales, which are discrete and cannot be backpropagated. GraphSim \cite{graphsim} adopts a breadth-first search (BFS)-based ordering scheme to capture structural information, but it can introduce permutation sensitivity and yield unstable performance. Additionally, GED and MCS similarities are typically learned separately in different training processes due to their distinct objectives. In contrast, our proposed \textsc{G2R} is an efficient model that effectively leverages the scale and structural information. Additionally, \textsc{G2R} is able to compute MCS and GED similarities simultaneously, offering a more holistic estimation of graph similarity.

\textbf{Geometric Representation Learning.} 
Geometric representation learning involves capturing the hierarchical relationships in embedding spaces, where entities are often modeled using shapes like boxes \cite{vilnis2018probabilistic, Li2018SmoothingTG, dasgupta2020improving}, cones \cite{Vendrov2015OrderEmbeddingsOI, Ganea2018HyperbolicEC} or disks \cite{Suzuki2019HyperbolicDE}, etc. They have found applications in computer vision \cite{Vendrov2015OrderEmbeddingsOI}, neural language processing \cite{vilnis2018probabilistic, Suzuki2019HyperbolicDE} and knowledge graph \cite{Ren2020Query2boxRO}. Our work is related to box embeddings, which use hyperrectangles to model hierarchical and disjoint relationships. Each hyperrectangle represents an entity, with its left lower and right upper corners defining the boundaries. The volume of the box serves as a measure of probability. 
We extend box embedding techniques by incorporating graph structural information and utilizing volume to represent graph size, capturing the intricate relationships and connectivity patterns. Unlike prior methods focusing solely on hierarchical modeling, our approach combines both hierarchical and structural information.

\section{Preliminaries} \label{sec:pre}
\textbf{Notation.} Let $G = (\mathcal{V}, \mathcal{E})$ be an undirected and connected graph, where each node $v \in \mathcal{V}$ is interconnected by $e(\cdot,\cdot)\in \mathcal{E}$. $|\mathcal{V}|$ denotes the number of nodes, while $|\mathcal{G}|$ denotes the cardinality of both nodes and edges.

\textbf{Subgraph Isomorphism.} A graph $M$ is isomorphic to a subgraph $G^\prime$ of $G$, if and only if there exists a \emph{bijective} function $f: \mathcal{V}_{M} \mapsto \mathcal{V}_{G^\prime}$, such that (1) $\forall v \in \mathcal{V}_{M}$ and $\forall f(v) \in \mathcal{V}_{G^\prime}$ have the same label if exists; (2) $\forall e(v,v^\prime)\in \mathcal{E}_{M}$, there exists $e(f(v),f(v^\prime))\in \mathcal{E}_{G^\prime}$. The function $f$ is called a \emph{mapping}. The pair $(M,G^\prime)$ represents a \emph{graph isomorphism}, while $(M,G)$ represents a \emph{subgraph isomorphism}.

\textbf{Maximum Common Subgraph (MCS).} The maximum common graph between $G_1$ and $G_2$ is $M$, if and only if there exists \emph{subgraph isomorphism} from $M$ to $G_1$ and $M$ to $G_2$, and $M$ has the most nodes compared with other common subgraphs. Such $M$ can be noted as $MCS(G_1, G_2)$.

\textbf{Graph Edit Distance (GED).} Given $G_1$ and $G_2$, their edit distance $GED(G_1, G_2)$ is the minimum cost of the sequence of edit operations that transform $G_1$ into a graph that is \emph{isomorphic} to $G_2$. The allowed operations are node/edge inserting, deleting, and label substituting.

MCS and GED are connected by Bunke in \cite{BUNKE1997689}, which suggests that, when label substitution is disabled, and the cost of removing or inserting an edge incident to a removed or inserted node is zero, computing the GED is equivalent to identifying the MCS. We refer to this GED as Bunke Graph Edit Distance (Bunke GED):

\textbf{Bunke Graph Edit Distance.}\label{def:bunkeGED}  Let $GED_{Bunke}$ denote the graph edit distance under the above specific cost function. It can be written as follows: $GED_{Bunke}(G_1, G_2) = |\mathcal{V}_{G_1}|+ |\mathcal{V}_{G_2}|-2\cdot|\mathcal{V}_M|$.

\textbf{Region.} A ``region" refers to an axis-aligned hyperrectangle $\in \mathbb{R}^d$, where each dimension is defined by a minimum value (the lower bound) and a maximum value (the upper bound).

Compared with box embedding \cite{vilnis2018probabilistic}, a ``region" remains invariant in volume and shape but allows for flexible adjustments in the left lower and right upper bounds, i.e., the shift of the hyperrectangle.

\section{Concurrent Computation of MCS and GED: Motivation and Analysis} \label{sec:investigation}
\subsection{Motivation}
As defined in the previous section, both MCS and GED are rooted in the concepts of isomorphism and subgraph isomorphism. While MCS focuses on local structural similarities by identifying the largest common subgraph, GED quantifies the global dissimilarity by considering the minimum number of transformations required to align the graphs. They provide complementary perspectives on graph similarity, offering insights into both commonalities and differences between graphs. By predicting both MCS and GED similarities concurrently, we can incorporate both local and global structural information, ensuring a more holistic assessment of graph similarity. To this end, we analyze the relationship between MCS and GED to explore their concurrent computation. In the following, we first introduce pairwise graph union to gain a holistic view of MCS, Bunke GED, and GED.

\subsection{Analysis}
Considering the \emph{binary graph union}, which merges the nodes and edges of two graphs based on their MCS. In this context, \emph{the disjoint parts} of the graphs refer to the nodes and edges not included in their MCS. Based on these, we present the following proposition: 

\textbf{Proposition 1.}  \emph{Given two graphs $G_1$ and $G_2$ and their MCS $M$, their graph union is $G = (\mathcal{V}_G, \mathcal{E}_G)$, where $|\mathcal{V}_G| = |\mathcal{V}_{G_1}| + |\mathcal{V}_{G_2}| - |\mathcal{V}_M|$ and $|\mathcal{E}_G| = |\mathcal{E}_{G_1}| + |\mathcal{E}_{G_2}| - |\mathcal{E}_M|$. Considering $\mathcal{V}_G$ as the universal set, $\mathcal{V}_M$ can be seen as the complement of the nodes in the disjoint parts. The number in these parts equals the Bunke GED.} Specifically, $|\mathcal{V}_M| = |\mathcal{V}_G| - GED_{\text{Bunke}}(G_1, G_2)$.

As Bunke GED considers only nodes, while GED involves both nodes and edges, we bridge the gap between them with \emph{Proposition 1} as follows: 
\begin{align}
|\mathcal{G}| - |\mathcal{M}|
&= |\mathcal{V}_G| - |\mathcal{V}_M| + |\mathcal{E}_{G}| - |\mathcal{E}_M| \notag\\
&= GED_{Bunke}(G_1, G_2) + |\mathcal{E}_{G_1}| + |\mathcal{E}_{G_2}| - 2 \cdot |\mathcal{E}_{M}|\notag\\
&= GED_{Bunke}(G_1, G_2) +\Phi(G_1,G_2) \label{eq:e1}\\
&\mbox{where } \Phi(G_1,G_2) =  |\mathcal{E}_{G_1}| + |\mathcal{E}_{G_2}| - 2 \cdot |\mathcal{E}_{M}| \in \mathbb{N}\label{eq:e2}
\end{align}
The above Eq. (\ref{eq:e2}) is derived under the existence of subgraph isomorphism from $M$ to $G_1$ and $G_2$, which implies that $0\leq|\mathcal{E}_M| \leq \operatorname{min}(|\mathcal{E}_{G_1}|, |\mathcal{E}_{G_2}|)$. 

Inheriting the structural constraint from isomorphism, GED should capture the dissimilarity between graphs while accounting for their shared substructures. Suppose that minimizing the GED between two graphs requires preserving all their shared substructures, which leads to the following proposition:

\textbf{Proposition 2.} \emph{Let us denote the union of common substructures between $G_1$ and $G_2$ excluding the MCS as $C = (\mathcal{V}_C, \mathcal{E}_C)$. It is important to note that $C$ may consist of multiple disconnected subgraphs. The cardinality of the node set $|\mathcal{V}_C|$ satisfies $0 \leq |\mathcal{V}_C| \leq \min(|\mathcal{V}_{G_1}|, |\mathcal{V}_{G_2}|) - |\mathcal{V}_M|$, and similarly, $|\mathcal{E}_C|$ satisfies $0 \leq |\mathcal{E}_C| \leq \min(|\mathcal{E}_{G_1}|, |\mathcal{E}_{G_2}|) - |\mathcal{E}_M|$.
In light of this, we can establish an upper bound for the GED as follows:
\begin{align}
GED(G_1, G_2) &\leq |\mathcal{G}| - |\mathcal{M}| - |\mathcal{C}| \notag\\
&\leq GED_{Bunke}(G_1, G_2) + \Phi(G_1,G_2) - |\mathcal{V}_C| - |\mathcal{E}_C| \notag
\end{align}}

\begin{figure*}[!h]
    \centering
    \includegraphics[width=\linewidth]{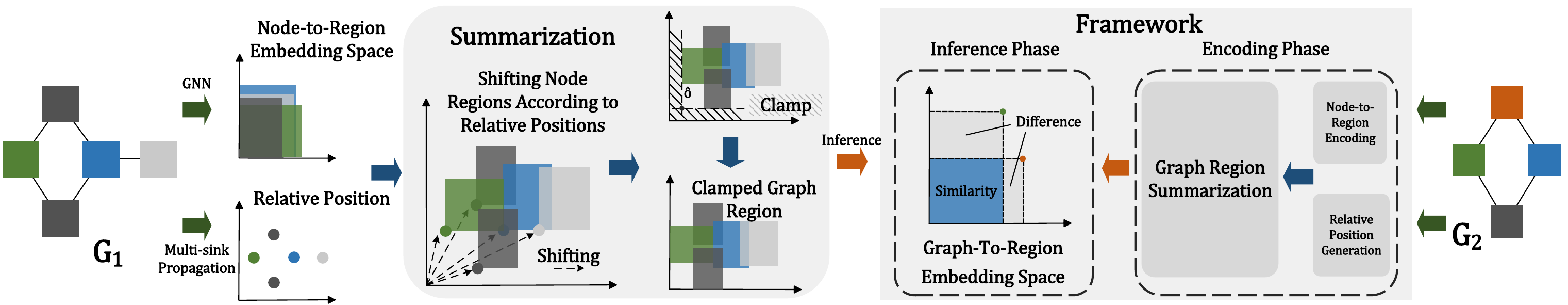}
    \caption{Overview of G2R. Given a graph as input, G2R projects each node onto the \emph{Node-To-Region Embedding Space} using a GNN and calculates nodes' relative positions through \emph{Multi-sink Propagation} to reflect their adjacency pattern in the embedding space. G2R then shifts the node regions from the global ordinate origin to their relative positions. Based on the shifted node regions, G2R summarizes the graph region and “re-shifts" it back to the origin. During inference, given the graph regions of two graphs as input, G2R predicts their MCS and GED similarities based on their overlapped and disjoint regions, respectively.}
    \label{fig:overview}
\end{figure*}

We focus on the case where $|\mathcal{V}_C| = 0$ and $|\mathcal{E}_C| = 0$, which leads to a looser upper bound for $GED(G_1, G_2)$. This upper bound provides a more flexible estimate and accounts for uncertainties. We can rewrite it as follows:
\begin{align}
GED(G_1, G_2) &\leq GED_{Bunke}(G_1, G_2) + \Phi(G_1,G_2) - 0 \notag\\
&\leq (1+\frac{\Phi(G_1,G_2)}{GED_{Bunke}(G_1, G_2)})\cdot \notag\\&\quad \quad \quad \quad \quad \quad \quad GED_{Bunke}(G_1, G_2)\notag\\
&\leq (1+\frac{|\mathcal{E}_{G_1}| + |\mathcal{E}_{G_2}| - 2 \cdot |\mathcal{E}_{M}|}{|\mathcal{V}_{G_1}|+ |\mathcal{V}_{G_2}|-2|\mathcal{V}_M|})\cdot \notag\\&\quad \quad \quad \quad \quad \quad \quad GED_{Bunke}(G_1, G_2)\label{eq:e4}\\
&\mbox{where \space} \frac{|\mathcal{E}_{G_1}| + |\mathcal{E}_{G_2}| - 2 \cdot |\mathcal{E}_{M}|}{|\mathcal{V}_{G_1}|+ |\mathcal{V}_{G_2}|-2|\mathcal{V}_M|} > 0\notag\\&\quad \quad \quad \quad \quad \quad \quad \mbox{ and }  |\mathcal{E}|\propto|\mathcal{V}|\label{eq:e5}
\end{align}

Specifically, Eq. (\ref{eq:e5}) holds due to the positive correlation between the number of edges and nodes in each graph. Additionally, Eq. (\ref{eq:e4}) highlights \emph{the positive correlation between Bunke GED and an upper bound of GED}, which further connects GED and MCS. 

\textbf{Summary.} Our exploration highlights that (1) Given $G_1$ and $G_2$, their Bunke GED equals the number of nodes in their disjoint parts; (2) Bunke GED can serve as a proxy for GED approximation as it exhibits a positive correlation with an upper bound of GED.

\section{Proposed Model}
In this section, we present \textsc{Graph2Region} (\textsc{G2R}), a geometric-based graph embedding model. It explicitly preserves structural and scale information of graphs for efficient graph similarity score computation. As illustrated in Fig. \ref{fig:overview}, \textsc{G2R} comprises two phases: encoding and inference. In the encoding phase, \textsc{G2R} transforms the input graph into graph regions within a solution space (Sec. \ref{subsec:model}), where region shapes capture the underlying graph structures. During the inference phase, \textsc{G2R} constrains the volume of regions to further reflect the graph scale. Leveraging our investigative findings in Sec. \ref{sec:investigation}, \textsc{G2R} computes the MCS and GED similarities of graph pairs based on their overlapped regions and disjoint parts, respectively. This decoupling of input for predictor grants our model the unique power of simultaneously computing MCS and GED similarities (Sec. \ref{subsec:inference}).

\subsection{Encoding Phase: Graph Region Modeling} \label{subsec:model}
Our key concept is representing graphs as regions, preserving their structural and scale properties in an embedding space. This procedure involves three stages: (1) \emph{Node-to-Region Encoding}: extracting neighborhood information from nodes and projecting it onto an embedding space; (2) \emph{Relative Position Generation}: computing relative positional encoding for each node, reflecting their proximity within graphs via a \emph{Multi-sink Propagation} mechanism; (3) \emph{Graph Region Summarization}: augmenting node regions with their relative positions to generate graph regions.

\subsubsection{Node-to-Region Encoding} Based on the Weisfeiler-Leman (WL) graph isomorphism test \cite{mckay2014practical}, the Message-Passing Graph Neural Networks (MPNNs) \cite{kipf2016semi, Velickovic2017GraphAN, Xu2018HowPA} takes a bag of node features $\mathbf{X}$ with a size of $|\mathcal{V}|\times d$ as input. Through WL-subtree guided message aggregation, MPNNs update the representations of each node based on their neighborhood. Consequently, MPNNs generate identical embeddings for nodes sharing the same WL subtree. We leverage this property of MPNNs to ensure that nodes with identical neighborhoods generate equivalent regions.

To convert discrete node labels into continuous values, we employ a linear layer and subsequently use a $k$-layer MPNN as the Node-to-Region encoder. We define the procedure for extracting information for node $v$ at layer $l$ as follows: 
\begin{equation}
    \mathbf{x}^{l}_v = \operatorname{UPDATE}(\mathbf{x}_v^{(l-1)},  \operatorname{AGGREGATE}(\mathbf{x}_u^{(l-1)}:{u\in\mathcal{N}(v)})\label{eq:feat_prop}
\end{equation}

Where $\mathbf{x} \in \mathbb{R}^d$, $u \in \mathcal{N}(u)$ refer to the neighbors of node $v$, $\operatorname{UPDATE}(\cdot)$ is a learnable function, $\operatorname{AGGREGATE}(\cdot)$ is a permutation-invariant operation, typically using max, mean, or sum. After the extracting procedure, we obtain a multi-scale representation $\mathbf{x} = \{\mathbf{x}^1,\dots,\mathbf{x}^k\}$ for each node. We then use a Multi-Layer Perceptron (MLP) to transform this representation into the corresponding multi-scale node region $\mathbf{r} = \{\mathbf{r}^1,\dots, \mathbf{r}^k\}$ where $\mathbf{r}^l\in \mathbb{R}^{D}$. We summarize this stage as $\phi(\mathbf{X}, \mathcal{A}) = \operatorname{MLP}_{e}(\operatorname{MPNN}(\operatorname{Linear}(\mathbf{X}),\mathcal{A}))$, where $\mathcal{A}$ represents the adjacency matrix of the graphs. We refer to it as $\phi(\mathbf{X})$ for short.

\begin{figure}[tb]
\centering
\includegraphics[width=\linewidth]{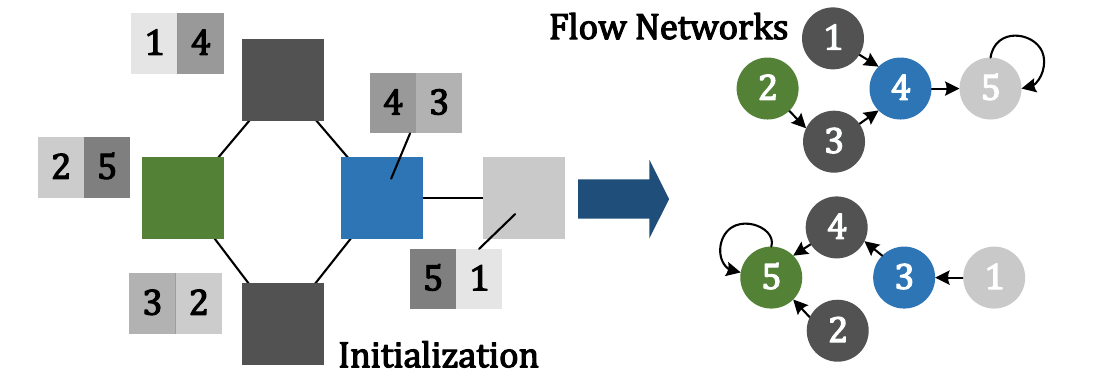}
\caption{Each node is assigned two random numbers (Left) to establish two flow networks (Right). After $3$ steps of propagation, the concatenated sequence for the blue node is [4,5,5,3,4,5], while for the light gray one is [5,5,5,1,3,4], compared with the green one [2,3,4,5,5,5], the blue node shared a similar sequence with the light gray one.} \label{fig:prop}
\end{figure}

\subsubsection{Relative Position Generation} The goal of this stage is to capture the structural information of graphs by encoding node proximity. To this end, we design a novel mechanism called \emph{Multi-sink Propagation}. This mechanism assign each node in the input graph a distinct random number, and direct its flow solely toward its neighbor with the highest value, establishing a flow network. Within this network, the node with the highest assigned value is the sink. The sink emerges as a focal point, toward which other nodes propagate through edges. Consequently, nodes nearby exhibit similar flow paths, capturing their proximity within the origin graph. However, flow paths can vary since different number assignments lead to diverse flow networks. To address this problem and capture inherent adjacency patterns, we alter the number assignment, allowing the computation of flow paths for each node toward multiple sinks.

\textsc{G2R} efficiently execute the propagation process using streamlined sparse scatter operations, similar to those in MPNN. To establish $d$ flow networks, \textsc{G2R} assign each node  in the graph $d$ continuous random numbers as its initial value, denoted as $\mathbf{s}^0 \in \mathbb{R}^{d}$. For $k$ steps of propagation, the paths for node $v$ at step $l$ can be formalized as follows: 
\begin{equation}
    \textbf{s}^l_v = \operatorname{MAX}(\textbf{s}_u^{l-1}: u\in\mathcal{N}(v))\label{eq:sink_prop}
\end{equation}
Where $\textbf{s}_v^l \in \mathbb{R}^d$ is landing points of $v$ at step $l$, $\textbf{s}_u^{l-1}$ is the possible landings points, $\operatorname{MAX(\cdot)}$ computes the dimension-wise maximum. At the end of the propagation process, we obtain  $\textbf{s} = [s^1,\dots,s^k] \in \mathbb{R}^ {k\times d}$ . We transport s into $\textbf{s}^\top \in \mathbb{R}^{d\times k}$ to obtain flow paths. 
We concatenate the paths for each node into a single sequence $\mathbf{S}\in \mathbb{R}^{d\cdot k}$, then pass it through an MLP to generate relative positions $\mathbf{o} = \operatorname{MLP}_{pe}(\mathbf{S})$, where $\mathbf{o} \in \mathbb{R}^D$ has the same dimensionality as the node regions. A running instance is illustrated in Fig \ref{fig:prop}. In summary, we describe this process as $\psi(\mathcal{A}) = \operatorname{MLP}_{pe}(\operatorname{PROP}(\mathcal{A}))$, and note as $\psi(\mathcal{A})$ for conciseness.

The proposed \emph{Multi-sink Propagation} offers two key advantages. First, it enables nearby nodes to exhibit similar directed flow paths, capturing their proximity and interconnectedness. These flow paths imitate fixed-length shortest paths toward high-valued nodes, with some repetitions, offering a more deterministic exploration of graph structure compared to DeepWalk \cite{deepwalk} and node2vec \cite{node2vec}, which use stochastic traversal strategies. Second, our mechanism eliminates the need for pretraining. It uses inexpensive random numbers and the message-passing mechanism of GNNs to compute the flow paths, ensuring efficient GPU implementation. These flow paths that can be trained jointly with downstream tasks, enabling end-to-end prediction.

\subsubsection{Graph Region Summarization} This stage generates graph regions for downstream prediction while considering the structural information of graphs. For this purpose, we combine the node regions $\mathbf{r}$ obtained from $\phi(\mathbf{X})$ with the relative position $\mathbf{o}$ generated by $\psi(\mathcal{A})$. One sensible approach is to augment the node regions by adding the relative position: $\hat{\mathbf{r}} = \mathbf{r} + \mathbf{o}$, where $\hat{\mathbf{r}} = \{\hat{\mathbf{r}}^1,\dots,\hat{\mathbf{r}}^k\}$, $\hat{\mathbf{r}}^l\in\mathbb{R}^D$. We summarize each graph by applying a pooling operation to the augmented node regions across all dimensions: $\mathbf{R} = \operatorname{POOL}(\hat{\mathbf{r}})$. This yields multi-scale graph regions $\mathbf{R} = \{\mathbf{R}^1,\dots,\mathbf{R}^k\}$, with $\mathbf{R}^l\in \mathbb{R}^{D}$, which preserve the structural information of original graphs. The volume of graph regions is further constrained in the inference phase in Sec. \ref{subsec:inference} to reflect the scale of graphs.

However, the initial value $\mathbf{s}^0$ may affect the generation of relative positions, where nodes with the equivalent connectivity pattern but different initial numbers may be projected onto different positions in the embedding space. To address this concern, we calculate the left lower corner of each graph region as $\hat{\mathbf{o}} = \operatorname{MIN}(\mathbf{o})$, where $\operatorname{MIN}(\cdot)$ determines the dimension-wise minimum across the relative positions of each node within the graph. We then adjust the graph regions by \textbf{clamping} them with $\hat{\mathbf{o}}$: $\hat{\mathbf{R}} = \mathbf{R} - \hat{\mathbf{o}}$, effectively “re-shifting" the graph regions back to the origin.

Finally, we pass the clamped graph region through a linear layer, which projects them onto the solution space, denoted as $\tilde{\mathbf{R}} = \operatorname{Linear}(\hat{\mathbf{R}})$. This linear layer implicitly aligns the maximum shared substructure of graph pairs, representing it as the overlapped region in the embedding space. At the end of the encoding phase, we obtain the multi-scale graph region for each graph, denoted as $\tilde{\mathbf{R}} = \{\tilde{\mathbf{R}}^1,\dots,\tilde{\mathbf{R}}^k\}$, where $\tilde{\mathbf{R}}^l \in \mathbb{R}^{out}$, $out$ represents the output dimension of each graph region. The pseudo algorithm of the entire encoding phase of \textsc{G2R} is available in Algorithm \ref{algo:enc}.

\begin{algorithm2e}[!htb]
\caption{The Encoding Phase of \textsc{G2R}}\label{algo:enc}
\LinesNumbered
\SetKwInOut{Input}{Input}
\SetKwInOut{Output}{Output}
\SetKwComment{Comment}{\# }{}
\Input{A graph G = ($\mathbf{X}$, $\mathcal{A}$);}
\Output{ Multi-scale graph region $\tilde{\mathbf{R}}$;}
\BlankLine
Init: $\mathbf{x}^0 = \operatorname{Linear}_0(\mathbf{X})$;\\
Generate multi-scale node embeddings $\mathbf{x}_v\in\mathbb{R}^{k\times d}$ with GNN based on Eq. (\ref{eq:feat_prop});\\
Node region projection: $\mathbf{r} = \operatorname{MLP}_e(\mathbf{x})$;\\
Flow init: $\mathbf{s}^0$ = Random\_value\_assign($|\mathcal{V}_G|$, n);\\
Calculte flow paths towards sinks $\mathbf{S}_v\in\mathbb{R}^{nj}$ with Multi-sink Propagation based on Eq. (\ref{eq:sink_prop});\\
Relative origin generation: $\mathbf{o} = \operatorname{MLP}_{pe}(\mathbf{S})$; \\
Augmenting node regions: $\hat{\mathbf{r}} = \mathbf{r} + \mathbf{o}$;\\
Graph Region Summarization: $\mathbf{R} = \operatorname{POOL}(\hat{\mathbf{r}})$;\\
Left lower corner computation: $\hat{\mathbf{o}} = \operatorname{MIN}(\mathbf{o})$;\\
Clamping: $\hat{\mathbf{R}} = \mathbf{R} - \hat{\mathbf{o}}$;\\
Graph region projection: $\tilde{\mathbf{R}} = \operatorname{Linear}(\hat{\mathbf{R}})$;\\
\end{algorithm2e}

\subsection{Inference Phase: Similarity Computation} \label{subsec:inference}
In this section, \textsc{G2R} constrain the graph regions derived from the input graph pair to reflect the graph scale. It then predicts MCS and GED similarities based on the shape and volume of the overlapped and disjoint parts of these regions, respectively. To this end, we introduce geometric operators for calculating the overlap and disjoint between the graph regions.

\subsubsection{Geometric Operators} The geometric intersection operator determines the shape of the overlapped region between two graph regions, while the volume operator calculates the volume of the involved regions for further prediction:
\paragraph{Geometric Intersection}
    \[\operatorname{Inter}(\tilde{\mathbf{R}}_{G_1}^l,\tilde{\mathbf{R}}_{G_2}^l) = \operatorname{MIN}(\tilde{\mathbf{R}}_{G_1}^l,\tilde{\mathbf{R}}_{G_2}^l)\]\\
    Where $\operatorname{MIN}(\cdot)$ is the dimension-wise minimum, $l$ is the $l$-th layer, and $\operatorname{Inter}(\tilde{\mathbf{R}}_{G_1}^l,\tilde{\mathbf{R}}_{G_2}^l) \in \mathbb{R}^{out}$.
\paragraph{Geometric Volume}
    \[\operatorname{Vol}(\tilde{\mathbf{R}}_{G}^l) = \prod^{out}_{i=1} \tilde{\mathbf{R}}_{G}^l[i]\]\\
    Where $\operatorname{Vol}(\tilde{\mathbf{R}}_{G}^l) \in \mathbb{R}$ and $i$ indicates the $i$-th dimensionality.
\subsubsection{Overlap for MCS Similarity} Given the multi-scale graph regions $\tilde{\mathbf{R}}_{G_1}$ and $\tilde{\mathbf{R}}_{G_2}$ of the input graph pair, we combine two scores to predict MCS similarity: the shape score and the volume score. For the shape score, we calculate and concatenate the multi-scale intersection $\operatorname{Inter}(\tilde{\mathbf{R}}_{G_1}, \tilde{\mathbf{R}}_{G_2}) \in \mathbb{R}^{k\times out}$ of graph regions. We feed the concatenated intersection into an MLP as follows:
\begin{align}
\mathbf{Score}_s(\tilde{\mathbf{R}}_{G_1}, \tilde{\mathbf{R}}_{G_2}) &= \operatorname{MLP}_{MCS}\notag\\&(\operatorname{CONCAT}(\operatorname{Inter}(\tilde{\mathbf{R}}_{G_1}, \tilde{\mathbf{R}}_{G_2}))) \label{eq:shape}
\end{align}
where $\mathbf{Score}_s(\tilde{\mathbf{R}}_{G_1}, \tilde{\mathbf{R}}_{G_2}) \in \mathbb{R}$. We normalize this score with the average node size of the input graphs to maintain scale awareness:
\begin{equation}
\hat{\mathbf{Score}}_s(\tilde{\mathbf{R}}_{G_1}, \tilde{\mathbf{R}}_{G_2}) = \frac{\mathbf{Score}_s(\tilde{\mathbf{R}}_{G_1}, \tilde{\mathbf{R}}_{G_2})}{(|G_1| + |G_2|)/2} \label{eq:norm_shape}
\end{equation}
For volume score, we first average the multi-scale graph regions: $\bar{\mathbf{R}}_{G} = \operatorname{MEAN}(\tilde{\mathbf{R}}_{G})$, where $\operatorname{MEAN}(\cdot)$ calculate the dimension-wise average for multi-scale graph regions and $\bar{\mathbf{R}}_{G} \in \mathbb{R}^{out}$. We then compute the volume score as follows to impose constraints on the volume of regions, reflecting the graph scale information:
\begin{equation}
\hat{\mathbf{Score}}_v(\bar{\mathbf{R}}_{G_1}, \bar{\mathbf{R}}_{G_2}) = \frac{\operatorname{Vol}(\operatorname{Inter}(\bar{\mathbf{R}}_{G_1}, \bar{\mathbf{R}}_{G_2}))}{(\operatorname{Vol(\bar{\mathbf{R}}_{G_1})}+ \operatorname{Vol(\bar{\mathbf{R}}_{G_2})})/2} \notag
\end{equation}
Where $\hat{\mathbf{Score}}_v(\bar{\mathbf{R}}_{G_1}, \bar{\mathbf{R}}_{G_2}) \in \mathbb{R}$. Finally, we combine these two scores to predict the MCS similarity:
\begin{align}
\mathbf{Score}_{MCS}(G_1,G_2) &= \alpha_1 \cdot \hat{\mathbf{Score}}_s(\tilde{\mathbf{R}}_{G_1}, \tilde{\mathbf{R}}_{G_2}) + \\
&\quad\quad\quad\quad \beta_1 \cdot \hat{\mathbf{Score}}_v(\bar{\mathbf{R}}_{G_1}, \bar{\mathbf{R}}_{G_2}) \notag
\end{align}
Where $\alpha_1, \beta_1 \in \mathbb{R}$ are two learnable weights.

\subsubsection{Difference as Proxy for GED} Based on the findings of our investigation in Sec. \ref{sec:investigation}, given $G_1$ and $G_2$, their Bunke GED equals the number of nodes in their disjoint parts and is positively correlated with the GED. Therefore, the Bunke GED can serve as a proxy for approximating GED similarity. To obtain this approximation, we calculate the geometric difference of graph regions as follows:
\[\operatorname{Difference}(\tilde{\mathbf{R}}_{G_1},\tilde{\mathbf{R}}_{G_2}) = \tilde{\mathbf{R}}_{G_1} + \tilde{\mathbf{R}}_{G_2} - 2 \cdot \operatorname{Inter}(\tilde{\mathbf{R}}_{G_1},\tilde{\mathbf{R}}_{G_2})\notag\]
Where $\operatorname{Difference}(\tilde{\mathbf{R}}_{G_1},\tilde{\mathbf{R}}_{G_2}) \in \mathbb{R}^{k \times out}$. As in the computation of MCS similarity, we calculate shape and volume scores for approximating GED similarity. We pass the concatenated difference shapes through $\operatorname{MLP}_{GED}(\cdot)$ to obtain the shape score $\widehat{\mathbf{Score}}_s \in \mathbb{R}$ as in Eq. (\ref{eq:shape}) and normalized as in Eq. (\ref{eq:norm_shape}). Similarly, we compute the volume score of the disjoint parts as follows:
\begin{equation}
\widehat{\mathbf{Score}}_v(\bar{\mathbf{R}}_{G_1}, \bar{\mathbf{R}}_{G_2}) = \gamma \cdot \frac{\operatorname{Vol}(\operatorname{Difference}(\bar{\mathbf{R}}_{G_1}, \bar{\mathbf{R}}_{G_2}))}{(\operatorname{Vol(\bar{\mathbf{R}}_{G_1})}+ \operatorname{Vol(\bar{\mathbf{R}}_{G_2})})/2} \notag
\end{equation}
Where $\gamma$ is a learnable weight that imitates the positive correlation in Eq. (\ref{eq:e4}). Finally, \textsc{G2R} predicts GED similarity score as follows:
\begin{align}
\mathbf{Score}_{GED}(G_1,G_2) &= \alpha_2 \cdot \widehat{\mathbf{Score}}_s(\tilde{\mathbf{R}}_{G_1}, \tilde{\mathbf{R}}_{G_2}) + \\&\quad\quad\quad\quad \beta_2 \cdot \operatorname{exp}(-\widehat{\mathbf{Score}}_v(\bar{\mathbf{R}}_{G_1}, \bar{\mathbf{R}}_{G_2}) )\notag
\end{align}
Where $\alpha_2, \beta_2 \in \mathbb{R}$ are learnable weights, and the exponential function $\operatorname{exp}(-x) = e^{-x}$ normalizes the GED similarity to the range $(0,1]$. We present pseudo-code of the inference stage in Algorithm \ref{algo:inf}. 
\subsubsection{Complexity} The complexity of generating node and graph regions is \(\mathcal{O}(|\mathcal{E}|)\), where \(|\mathcal{E}|\) is the number of edges in graphs. Similarly, the complexity of the \emph{Multi-sink Propagation} mechanism is also \(\mathcal{O}(|\mathcal{E}|)\), resulting in a total encoding phase complexity of \(\mathcal{O}(2\cdot|\mathcal{E}|)\). It is worth noting that the generation of graph regions relies solely on the original input graph, allowing this procedure to be preprocessed. The complexity of the inference phase is \(\mathcal{O}(out)\), which is significantly more efficient compared with other pairwise comparison models \cite{Bai2019SimGNNAN, graphsim, Doan2021InterpretableGS,  Roy2022MaximumCS} whose complexity is at least $\mathcal{O}(max(|\mathcal{V}_{G_1}|, |\mathcal{V}_{G_2}|)^2 \cdot out)$.

\begin{algorithm2e}[!htb]
\caption{The Inference Phase of \textsc{G2R}}\label{algo:inf}
\LinesNumbered
\SetKwInOut{Input}{Input}
\SetKwInOut{Output}{Output}
\SetKwComment{Comment}{\# }{}
\Input{The graph regions of a graph pair $\tilde{\mathbf{R}}_1, \tilde{\mathbf{R}}_2$;}
\Output{Predicted similarity score Score;}
\BlankLine
Calculate overlap of multi-scale graph regions $Inter\in \mathbb{R}^{kd}$;\\
Calculate the mean of graph regions $\bar{\mathbf{R}}_1$,$\bar{\mathbf{R}}_2\in\mathbb{R}^d$ and their overlap: $Inter_{mean} \in \mathbb{R}^d$;\\
\eIf{calulate MCS similarity}{
$Score_s = \operatorname{MLP}_{MCS}(Inter)$;\\
$Score_v = \frac{\operatorname{Vol}(Inter_{mean})}{((\operatorname{Vol}(\bar{\mathbf{R}}_1) + \operatorname{Vol}(\bar{\mathbf{R}}_2))/2)}$;\\
Score = $\alpha_1 * Score_s + \beta_1 * Score_v$; 
}{
Differ = $\tilde{\mathbf{R}}_1 + \tilde{\mathbf{R}}_2 - 2 * Inter$;\\ 
$Score_s = \operatorname{MLP}_{GED}$(Differ);\\
Differ$_{mean}$ = $\bar{\mathbf{R}}_1 + \bar{\mathbf{R}}_2 - 2 * Inter_{mean}$;\\
$Score_v = \gamma * \frac{\operatorname{Vol}(\operatorname{Differ}_{mean})}{((\operatorname{Vol}(\bar{\mathbf{R}}_1) + \operatorname{Vol}(\bar{\mathbf{R}}_2))/2)}$;\\
Score = $\alpha_2 * Score_s + \beta_2 * Score_v$; 
}
\end{algorithm2e}

\section{Model Training}
\subsubsection{Training Objective} Following previous works \cite{Doan2021InterpretableGS, zhuo2022efficient}, the training target for MCS and GED similarities are:
\begin{align}
    \operatorname{nMCS}(G_1, G_2) = \frac{|\mathcal{V}_{MCS(G_1, G_2)}|}{(|\mathcal{V}_{G_1}| + |\mathcal{V}_{G_2}|)/2}\\
    \operatorname{nGED}(G_1, G_2) = \frac{GED(G_1, G_2)}{(|\mathcal{V}_{G_1}| + |\mathcal{V}_{G_2}|)/2}
\end{align}
The $\operatorname{nGED}(\cdot,\cdot)$ value is normalized using the exponential function $\operatorname{exp}(-x) = e^{-x}$.

We train models with the \emph{Mean Squared Error (MSE) Loss}, minimizing the predictive errors between the predicted scores and normalized ground truths:
\begin{align}
    \mathcal{L}_{t} = \frac{1}{|\mathcal{N}|}&\sum_{(i,j)\in\mathcal{N}}(\mathbf{Score}_{t}(G_i,G_j) - \widetilde{\mathbf{Score}}_{t}(G_i,G_j))^2\notag\\
    &\quad\quad\quad\quad\quad\quad  \mbox{where \space} t\in\{\operatorname{MCS},\operatorname{GED}\}
\end{align}
Where $\mathcal{N}$ is the collection of graph pairs, $\widetilde{\mathbf{Score}}_{t}(G_i,G_j)$ is the normalized ground truth.

\subsubsection{One Train To Predict Them All} Our model decouples the input for downstream similarity prediction. It predicts MCS similarity using the overlapped region, and approximates GED similarity based on the disjoint regions of graph pairs. To our knowledge, our model is the first approach to simultaneously predict MCS and GED similarities. The simplest dual training objective is given by:
\[\mathcal{L}_{dual} = \mathcal{L}_{MCS} + \mathcal{L}_{GED}\]
However, joint MCS and GED similarity computations can be regarded as a multi-task learning problem. To prioritize efficiency, we use the uncertainty-weighted loss \cite{uncertainty}, one of the most efficient and impactful multi-task learning losses. This loss dynamically learns task uncertainty from predictive errors. Unlike other multi-task learning losses, it ensures stability while requiring no additional supervision or external priors, and has shown competitive performance in many computer vision tasks. The uncertainty-weighted objective is given by:
\[\mathcal{L}_{dual}^{+} = \sum_{t \in \{MCS, GED\}} \frac{1}{2} \cdot(\operatorname{exp}(-\theta_{t}) \cdot \mathcal{L}_{t} + \theta_{t})\]
Where $\theta_{t} \in \mathbb{R}$ is two learnable weights. The \textsc{G2R} model that incorporates both MCS and GED similarities as training targets, we refer to as \textsc{G2R-Dual}.

\section{Evaluations}
In this section, we compare \textsc{G2R} against 9 baselines over 16 datasets in terms of (1) \emph{Effectiveness}: we evaluated the effectiveness of our model compared with state-of-the-art models in MCS similarity learning; (2) \emph{Transferability}: we trained the models on smaller synthetic graphs, then assessed their transferability on larger, unseen real-world graphs; (3) \emph{Concurrent Prediction}: we evaluated the performance of $\textsc{G2R}$ and $\textsc{G2R-Dual}$ in MCS and GED similarity computations; (4) \emph{Ablation Study} we analyzed the key components of $\textsc{G2R}$ and $\textsc{G2R-Dual}$. (5) \emph{Time Efficiency}: we evaluated the time efficiency of models in terms of training and inference times as well as the convergence speed. (6) \emph{Hyperparameter Sensitivity}: we analyzed the influence of flow paths and GNN layers; (7) \emph{Interpretability}: we analyzed the parameter learned of our trained model; (8) \emph{Case Study}: we visually analyzed the ranking results of \textsc{G2R} and the pairwise node comparisons to gain a deeper understanding of the \textsc{G2R}'s behavior.

\subsection{Experimental Setup}
\begin{table}[h]
    \centering
    \caption{Statistics of datasets used in MCS similarity learning. $D$ means the diameter. $-$ means unlabeled.}
    \resizebox{\linewidth}{!}{
    \begin{tabular}{l|cccccccccc}
    \toprule
        \textbf{Dataset} &
        \# Graphs &
        \# Features &
        Avg. $|\mathcal{V}_G|$ &
        Avg. $|\mathcal{E}_G|$ &
        Max. $|\mathcal{V}_G|$ &
        Max. $|\mathcal{E}_G|$ &
        Avg. D &
        Max. D &
        Min. D &
        \# Pairs \\
    \midrule
    \textsc{AIDS}  & 1955 & 38 & 12.7 & 12.6 & 62 & 67 & 7.3 & 33 & 2 & 3.82 M \\
    \textsc{COX2} & 2000 & 35 & 17.5 & 17.7 & 39 & 42 & 7.9 & 16 & 3 & 4 M \\
    \textsc{Enzymes} & 1997 & 3 & 17.11 & 28.39 & 80 & 89 & 6.8 & 33 & 1 & 3.99 M \\
    \textsc{IMDB-Binary} & 2000 & - & 11.7 & 42.0 & 93 & 804 & 1.8 & 3 & 1 & 4 M \\
    \textsc{MUTAG} & 2000 & 7 & 10.0 & 10.0 & 19 & 20 & 5.7 & 12 & 3 & 4 M \\
    \textsc{PTC-FM} & 1847 & 18 & 10.4 & 10.0 & 43 & 47 & 6.2 & 23 & 2 & 3.41 M \\
    \textsc{PTC-FR} & 1874 & 19 & 10.0 & 9.7 & 39 & 42 & 6.0 & 26 & 2 & 3.51 M \\
    \textsc{PTC-MM} & 1837 & 20 & 10.8 & 10.4 & 44 & 49 & 6.4 & 21 & 2 & 3.38 M \\
    \textsc{PTC-MR} & 1877 & 18 & 10.3 & 9.7 & 39 & 42 & 6.1 & 25 & 2 & 3.52 M \\
    \bottomrule
    
    \end{tabular}
    }
    \label{tab:stats_mcs}
\end{table}

\subsubsection{Datasets} We evaluated our model over 15 datasets: (i) For MCS similarity learning, we selected 9 datasets \cite{Morris+2020} from small molecules ({AIDS}, {COX2}, {MUTAG}, {PTC-FM}, {PTC-FR}, {PTC-MM}, {PTC-MR}), bioinformatics ({Enzymes}) and social networks ({IMDB-Binary}), respectively. The exact MCS is computed using the \emph{McSplit} algorithm \cite{mcsplit}. For the statistics of the sampled graphs per dataset, please refer to Table \ref{tab:stats_mcs}. (ii) For MCS and GED similarities, we followed the experimental setup in  \cite{graphsim, Bai2019SimGNNAN, Doan2021InterpretableGS, zhuo2022efficient} and conducted experiments on three commonly used datasets: {AIDS700}, {Linux}, {IMDB-Multi}, with the true GED values provided in \cite{Bai2019SimGNNAN}, and the exact MCS calculated using \emph{McSplit}. (iii) For transferability and time efficiency, we trained models on synthetic ER graphs with $n\in[5,50], p_e\in[0.1,0.5]$, where $n$ is the node size and $p_e$ is the edge probability. We then evaluated the models on 4 real-world datasets \cite{Morris+2020,wordnet}: {MRSC\_21}, {D\&D}, {FirstMM\_DB} and {WordNet}, which contain larger graphs.

\subsubsection{Evaluation Metrics} 
We adopt \emph{Mean Square Error (MSE)} and \emph{Mean Absolute Error (MAE)} to evaluate the effectiveness of models in similarity learning, and \emph{Spearman's Rank Correlation Coefficient ($\rho$)}, \emph{Kendall's Rank Correlation Coefficient ($\tau$)} and \emph{Precision at $k$ ($p@k$)} for ranking. The average and standard deviation of the results from five different runs are reported.

\begin{table*}[]
\caption{Prediction and ranking of MCS similarity on test sets. The MSE and MAE are in $10^{-3}$. The best is highlighted in bold, while the second is underlined. Results marked with $\dag$ mean we report the best five runs among at least ten.}\label{tab:mcs}
\resizebox{\linewidth}{!}{
\begin{tabular}{l|lccccccccc|c}
\toprule
\multicolumn{2}{l}{\textbf{Dataset}}& 
\textsc{SimGNN} \cite{Bai2019SimGNNAN} &
\textsc{GraphSim} \cite{graphsim}&
\textsc{GMN} \cite{Li2019GraphMN}&
\textsc{GOTSim} \cite{Doan2021InterpretableGS}&
\textsc{LMCCS} \cite{Roy2022MaximumCS}&
\textsc{XMCS} \cite{Roy2022MaximumCS}&
\textsc{ERIC} \cite{zhuo2022efficient}&
\textsc{GEN} \cite{Li2019GraphMN}&
\textsc{Greed} \cite{ranjan2022greed}&
\textsc{G2R} (Ours)\\
\midrule
\multirow{5}{*}{\textsc{AIDS}} &
MSE $\downarrow$ &
$1.21 \pm 0.04$ &       
0.76 $\pm$ $0.04^\dag$ &
3.10 $\pm$ 0.21 &
7.49 $\pm$ 0.11 &
8.04 $\pm$ 0.45 &
0.99 $\pm$ 0.06 &
\underline{0.67 $\pm$ 0.08} &
3.53 $\pm$ 0.13 &
1.38 $\pm$ $0.11^\dag$ &
\textbf{0.30 $\pm$ 0.01} \\
& MAE $\downarrow$ &
23.46 $\pm$ 0.38 &    
15.48 $\pm$ $0.83^\dag$ & 
34.78 $\pm$ 1.63 &
66.68 $\pm$ 0.54 &
69.56 $\pm$ 0.00 &
20.94 $\pm$ 0.47 &
\underline{12.64 $\pm$ 1.11} & 
36.69 $\pm$ 0.75 &
27.35 $\pm$ $1.37^\dag$ &
\textbf{8.49 $\pm$ 0.39} \\
& $\rho \uparrow$ &
0.96 $\pm$ 0.00 &
\underline{0.97 $\pm$ $0.00^\dag$} &
0.95 $\pm$ 0.00 &
0.88 $\pm$ 0.00 &
0.84 $\pm$ 0.00 &
0.96 $\pm$ 0.00 &
\underline{0.97 $\pm$ 0.00} &
0.96 $\pm$ 0.00 & 
0.96 $\pm$ $0.00^\dag$ &
\textbf{0.99 $\pm$ 0.00} \\
& $\tau \uparrow$ &
0.84 $\pm$ 0.00 &   
0.88 $\pm$ $0.00^\dag$ &
0.82 $\pm$ 0.01 &
0.73 $\pm$ 0.00 &
0.67 $\pm$ 0.01 &
0.85 $\pm$ 0.00 &
\underline{0.89 $\pm$ 0.00} &
0.84 $\pm$ 0.00 &
0.85 $\pm$ $0.01^\dag$ &
\textbf{0.91 $\pm$ 0.00} \\
& $p@10 \uparrow$ &
0.55 $\pm$ 0.02 &  
0.67 $\pm$ $0.01^\dag$ &
0.65 $\pm$ 0.01 &
0.17 $\pm$ 0.01 &
0.40 $\pm$ 0.03 &
0.58 $\pm$ 0.01 &
0.68 $\pm$ 0.02 &
\underline{0.69 $\pm$ 0.01} &
0.68 $\pm$ $0.02^\dag$ &
\textbf{0.77 $\pm$ 0.08} \\
\midrule
\multirow{5}{*}{\textsc{COX2}} & 
MSE $\downarrow$ &
1.19 $\pm$ 0.04 & 
1.36 $\pm$ $0.05^\dag$ &
1.75 $\pm$ 0.08 &
4.99 $\pm$ $0.59^\dag$ &
5.06 $\pm$ $0.30^\dag$ &
1.46 $\pm$ 0.07 &
\underline{0.65 $\pm$ 0.04} &
1.23 $\pm$ 0.05 &
0.89 $\pm$ $0.03^\dag$ &
\textbf{0.52 $\pm$ 0.02} \\
& MAE $\downarrow$ &
25.89 $\pm$ 0.48 &     
27.49 $\pm$ $0.76^\dag$ &
32.27 $\pm$ 0.84 &
55.15 $\pm$ $4.01^\dag$ &
56.25 $\pm$ $1.92^\dag$ &
28.75 $\pm$ 0.57 &
\underline{18.56 $\pm$ 0.62} &
25.62 $\pm$ 0.49 &
22.64 $\pm$ $0.40^\dag$ &
\textbf{15.78 $\pm$ 0.03} \\
& $\rho \uparrow$ &
0.93 $\pm$ 0.00 & 
0.92 $\pm$ $0.00^\dag$ &
0.92 $\pm$ 0.00 &
0.82 $\pm$ $0.01^\dag$ &
0.77 $\pm$ $0.01^\dag$ &
0.91 $\pm$ 0.01 &  
\textbf{0.96 $\pm$ 0.00} &
\underline{0.95 $\pm$ 0.00} &
\underline{0.95 $\pm$ $0.00^\dag$} &
\textbf{0.96 $\pm$ 0.00} \\
& $\tau \uparrow$ &
0.79 $\pm$ 0.00 &
0.78 $\pm$ $0.00^\dag$ &
0.77 $\pm$ 0.00 &
0.65 $\pm$ $0.01^\dag$ &
0.59 $\pm$ $0.01^\dag$ &
0.76 $\pm$ 0.01 &
\underline{0.84 $\pm$ 0.00} &
0.82 $\pm$ 0.00 &
0.83 $\pm$ $0.00^\dag$ &
\textbf{0.86 $\pm$ 0.00} \\
& $p@10 \uparrow$ &
0.29 $\pm$ 0.01 &
0.27 $\pm$ $0.01^\dag$ &
0.47 $\pm$ 0.01 &
0.04 $\pm$ $0.00^\dag$ &
0.16 $\pm$ $0.02^\dag$ &
0.19 $\pm$ 0.02 &
\underline{0.56 $\pm$ 0.02} &
\textbf{0.61 $\pm$ 0.01} &
0.53 $\pm$ $0.01^\dag$ &
\textbf{0.61 $\pm$ 0.01} \\
\midrule
\multirow{5}{*}{\textsc{Enzymes}} &
MSE $\downarrow$ &
2.14 $\pm$ 0.05 &   
2.93 $\pm$ $0.11^\dag$ &
7.10 $\pm$ 0.25 &
6.04 $\pm$ $0.02^\dag$ &
9.82 $\pm$ 1.67 &
2.15 $\pm$ 0.08 &
\underline{1.57 $\pm$ 0.05} &
6.73 $\pm$ 0.13 &
1.76 $\pm$ $0.03^\dag$ &
\textbf{1.33 $\pm$ 0.01} \\
& MAE $\downarrow$ &
35.39 $\pm$ 0.40 &
41.65 $\pm$ $0.86^\dag$ &
65.81 $\pm$ 1.42 & 
61.93 $\pm$ $0.13^\dag$ &
79.93 $\pm$ 7.13 &
34.96 $\pm$ 0.47 &
\underline{29.98 $\pm$ 0.36} &
63.26 $\pm$ 0.19 &
32.13 $\pm$ $0.31^\dag$ &
\textbf{27.67 $\pm$ 0.13} \\
& $\rho \uparrow$ &
0.89 $\pm$ 0.00 &
0.85 $\pm$ $0.01^\dag$ &
0.80 $\pm$ 0.01 &
0.81 $\pm$ $0.00^\dag$ &
0.50 $\pm$ 0.14 &
0.89 $\pm$ 0.01 &
\textbf{0.93 $\pm$ 0.00} &
0.82 $\pm$ 0.00 &
\underline{0.91 $\pm$ $0.00^\dag$} &
\textbf{0.93 $\pm$ 0.00} \\
& $\tau \uparrow$ &
0.72 $\pm$ 0.00 &
0.68 $\pm$ $0.01^\dag$ &
0.62 $\pm$ 0.01 &
0.64 $\pm$ $0.00^\dag$ &
0.36 $\pm$ 0.10 &
0.73 $\pm$ 0.01 &
\underline{0.77 $\pm$ 0.00} &
0.65 $\pm$ 0.00 &
0.76 $\pm$ $0.00^\dag$ &
\textbf{0.79 $\pm$ 0.00} \\
& $p@10 \uparrow$ &
0.13 $\pm$ 0.01 &
0.07 $\pm$ $0.01^\dag$ &
0.14 $\pm$ 0.01 &
0.06 $\pm$ $0.00^\dag$ &
0.03 $\pm$ 0.02 &
0.12 $\pm$ 0.01 &
\underline{0.20 $\pm$ 0.02} &
0.16 $\pm$ 0.05 &  
\underline{0.20 $\pm$ $0.01^\dag$} &
\textbf{0.25 $\pm$ 0.01} \\
\midrule
\multirow{5}{*}{\textsc{IMDB-Binary}} &
MSE $\downarrow$ &
0.79 $\pm$ 0.10 & 
1.52 $\pm$ $0.04^\dag$ &
0.64 $\pm$ 0.09 &
15.36 $\pm$ 0.06 &
3.62 $\pm$ $0.04^\dag$ &
1.89 $\pm$ 0.35 &
\underline{0.35 $\pm$ 0.04} &
0.56 $\pm$ 0.04 &
0.88 $\pm$ $0.03^\dag$ &
\textbf{0.31 $\pm$ 0.05} \\
& MAE $\downarrow$ &
20.00 $\pm$ 1.57 & 
23.35 $\pm$ $2.28^\dag$ &
16.93 $\pm$ 1.77 &
99.44 $\pm$ 0.37 &
46.47 $\pm$ $0.35^\dag$ &
32.83 $\pm$ 3.42 &
\underline{10.10 $\pm$ 0.79} &
16.12 $\pm$ 0.72 &
20.30 $\pm$ $0.61^\dag$ &
\textbf{8.90 $\pm$ 0.47} \\
& $\rho \uparrow$ &
\underline{0.97 $\pm$ 0.00} &
0.95 $\pm$ $0.00^\dag$ &
\textbf{0.98 $\pm$ 0.00} &
0.49 $\pm$ 0.01 &
0.90 $\pm$ $0.00^\dag$ &
0.93 $\pm$ 0.01 &
\textbf{0.98 $\pm$ 0.00} &
\textbf{0.98 $\pm$ 0.00} &
\underline{0.97 $\pm$ $0.00^\dag$} &
\textbf{0.98 $\pm$ 0.00} \\
& $\tau \uparrow$ &
0.87 $\pm$ 0.01 &
0.84 $\pm$ $0.00^\dag$ &
0.89 $\pm$ 0.01 &
0.36 $\pm$ 0.00 &
0.73 $\pm$ $0.00^\dag$ &
0.80 $\pm$ 0.02 &
\underline{0.91 $\pm$ 0.00} & 
0.90 $\pm$ 0.00 &
0.88 $\pm$ $0.00^\dag$ &
\textbf{0.92 $\pm$ 0.00} \\
& $p@10 \uparrow$ &
0.77 $\pm$ 0.02 &
0.69 $\pm$ $0.01^\dag$ &
0.83 $\pm$ 0.02 &
0.25 $\pm$ 0.03 &
0.36 $\pm$ $0.01^\dag$ &
0.54 $\pm$ 0.06 & 
0.88 $\pm$ 0.02 &
\textbf{0.90 $\pm$ 0.00} &
0.79 $\pm$ $0.01^\dag$ &
\underline{0.88 $\pm$ 0.01} \\
\midrule
\multirow{5}{*}{\textsc{MUTAG}} &
MSE $\downarrow$ &
1.59 $\pm$ 0.14 &
0.94 $\pm$ $0.05^\dag$ &
0.95 $\pm$ 0.06 & 
4.66 $\pm$ 0.07 &
5.53 $\pm$ 0.82 &
1.34 $\pm$ 0.06 &
\underline{0.48 $\pm$ 0.04} &
0.69 $\pm$ 0.03 &
6.24 $\pm$ $0.02^\dag$ &
\textbf{0.21 $\pm$ 0.02} \\
& MAE $\downarrow$ &
30.82 $\pm$ 1.48 &
18.06 $\pm$ $1.36^\dag$ &
23.03 $\pm$ 0.95 &
54.78 $\pm$ 0.44 &
58.85 $\pm$ 4.54 &
27.97 $\pm$ 0.52 &
\underline{11.81 $\pm$ 1.15} &
19.85 $\pm$ 0.37 &
61.40 $\pm$ $0.08^\dag$ &
\textbf{7.43 $\pm$ 0.59} \\
& $\rho \uparrow$ &
0.82 $\pm$ 0.01 &
0.89 $\pm$ $0.01^\dag$ &
0.91 $\pm$ 0.00 &
0.58 $\pm$ 0.00 &
0.41 $\pm$ 0.13 &
0.84 $\pm$ 0.01 &
\underline{0.94 $\pm$ 0.00} &  
0.93 $\pm$ 0.00 &
0.35 $\pm$ $0.00^\dag$ & 
\textbf{0.97 $\pm$ 0.00} \\
& $\tau \uparrow$ &
0.64 $\pm$ 0.02 &
0.74 $\pm$ $0.01^\dag$ &
0.75 $\pm$ 0.00 &
0.44 $\pm$ 0.00 &
0.29 $\pm$ 0.09 &
0.67 $\pm$ 0.01 &
\underline{0.81 $\pm$ 0.01} &
0.78 $\pm$ 0.00 &
0.25 $\pm$ $0.00^\dag$ &
\textbf{0.87 $\pm$ 0.01} \\
& $p@10 \uparrow$ &
0.46 $\pm$ 0.07 &       
0.55 $\pm$ $0.01^\dag$ &   
0.78 $\pm$ 0.00 &   
0.09 $\pm$ 0.01 &
0.19 $\pm$ 0.21 &  
0.51 $\pm$ 0.03 &    
0.77 $\pm$ 0.01 &     
\underline{0.81 $\pm$ 0.01} &    
0.02 $\pm$ $0.01^\dag$ &   
\textbf{0.84 $\pm$ 0.02} \\
\midrule
\multirow{5}{*}{\textsc{PTC-FM}} &
MSE $\downarrow$ &
1.50 $\pm$ 0.14 &  
1.44 $\pm$ $0.41^\dag$ &
1.43 $\pm$ 0.09 &
4.71 $\pm$ 0.05 &
7.26 $\pm$ 2.11 &
1.00 $\pm$ 0.10 &
\underline{0.70 $\pm$ 0.07} &   
1.05 $\pm$ 0.04 &
1.07 $\pm$ $0.03^\dag$ &  
\textbf{0.28 $\pm$ 0.02} \\
& MAE $\downarrow$ &
27.81 $\pm$ 1.50 &       
25.11 $\pm$ $6.70^\dag$ &   
27.50 $\pm$ 0.98 &   
54.99 $\pm$ 0.28 &
65.67 $\pm$ 9.77 &  
22.95 $\pm$ 1.28 & 
\underline{14.37 $\pm$ 0.80} &     
23.40 $\pm$ 0.40 &    
24.46 $\pm$ $0.31^\dag$ &    
\textbf{8.56 $\pm$ 0.24} \\
& $\rho \uparrow$ &
0.91 $\pm$ 0.01 &
0.91 $\pm$ $0.02^\dag$ &
0.93 $\pm$ 0.00 &
0.79 $\pm$ 0.00 &
0.64 $\pm$ 0.12 &
0.93 $\pm$ 0.01 &
\underline{0.95 $\pm$ 0.00} &
\underline{0.95 $\pm$ 0.00} &
0.94 $\pm$ $0.00^\dag$ &
\textbf{0.98 $\pm$ 0.00} \\
& $\tau \uparrow$ &
0.76 $\pm$ 0.01 &
0.77 $\pm$ $0.04^\dag$ &
0.79 $\pm$ 0.00 &
0.62 $\pm$ 0.00 &
0.49 $\pm$ 0.10 &
0.79 $\pm$ 0.01 &
\underline{0.84 $\pm$ 0.01} &
0.82 $\pm$ 0.00 &
0.81 $\pm$ $0.00^\dag$ &
\textbf{0.89 $\pm$ 0.00} \\
& $p@10 \uparrow$ &
0.47 $\pm$ 0.04 &
0.46 $\pm$ $0.11^\dag$ &
0.72 $\pm$ 0.01 &
0.13 $\pm$ 0.00 &
0.22 $\pm$ 0.17 &
0.63 $\pm$ 0.06 &
0.68 $\pm$ 0.03 &
\underline{0.78 $\pm$ 0.00} &
0.72 $\pm$ $0.00^\dag$ &
\textbf{0.82 $\pm$ 0.01} \\
\midrule
\multirow{5}{*}{\textsc{PTC-MR}} &
MSE $\downarrow$ &
1.50 $\pm$ 0.19 &
1.26 $\pm$ $0.32^\dag$ &
0.94 $\pm$ 0.13 &
5.01 $\pm$ 0.01 &
8.44 $\pm$ 2.59 &
0.87 $\pm$ 0.08 &
\underline{0.53 $\pm$ 0.05} &
0.98 $\pm$ 0.03 &
1.06 $\pm$ $0.05^\dag$ &
\textbf{0.26 $\pm$ 0.01} \\
& MAE $\downarrow$ &
27.97 $\pm$ 2.02 &
23.67 $\pm$ $5.68^\dag$ &
21.18 $\pm$ 1.73 &
57.23 $\pm$ 0.10 &
70.68 $\pm$ 10.96 &
21.23 $\pm$ 0.98 &
\underline{13.53 $\pm$ 0.79} &
22.25 $\pm$ 0.28 &
24.02 $\pm$ $0.73^\dag$ &
\textbf{7.98 $\pm$ 0.43} \\
& $\rho \uparrow$ &
0.89 $\pm$ 0.01 &
0.92 $\pm$ $0.01^\dag$ &
0.94 $\pm$ 0.01 &
0.77 $\pm$ 0.00 &
0.61 $\pm$ 0.14 &
0.93 $\pm$ 0.01 &
\underline{0.96 $\pm$ 0.00} &
0.95 $\pm$ 0.00 &
0.94 $\pm$ $0.00^\dag$ &
\textbf{0.98 $\pm$ 0.00} \\
& $\tau \uparrow$ &
0.74 $\pm$ 0.02 &
0.79 $\pm$ $0.02^\dag$ &
0.82 $\pm$ 0.01 &
0.61 $\pm$ 0.00 &
0.45 $\pm$ 0.11 &
0.80 $\pm$ 0.01 &
\underline{0.85 $\pm$ 0.01} &
0.82 $\pm$ 0.00 &
0.81 $\pm$ $0.00^\dag$ &
\textbf{0.89 $\pm$ 0.00} \\
& $p@10 \uparrow$ &
0.48 $\pm$ 0.06 &
0.54 $\pm$ $0.03^\dag$ &
0.76 $\pm$ 0.01 &
0.14 $\pm$ 0.01 &
0.25 $\pm$ 0.20 &
0.67 $\pm$ 0.05 &
0.74 $\pm$ 0.03 &
\underline{0.79 $\pm$ 0.00} &
0.75 $\pm$ $0.01^\dag$ &
\textbf{0.83 $\pm$ 0.01} \\
\midrule
\multirow{5}{*}{\textsc{PTC-MM}} &
MSE $\downarrow$ &
1.62 $\pm$ 0.15 &
1.41 $\pm$ $0.04^\dag$ &
1.45 $\pm$ 0.08 &
4.99 $\pm$ $0.11^\dag$ &
6.01 $\pm$ $0.25^\dag$ &
1.05 $\pm$ 0.08 &
\underline{0.65 $\pm$ 0.04} &
1.30 $\pm$ 0.04 &
1.18 $\pm$ $0.15^\dag$ &
\textbf{0.29 $\pm$ 0.01} \\
& MAE $\downarrow$ &
29.51 $\pm$ 1.41 &
27.05 $\pm$ $0.29^\dag$ &
26.96 $\pm$ 1.19 &
56.59 $\pm$ $0.84^\dag$ &
60.41 $\pm$ $1.51^\dag$&
23.56 $\pm$ 0.95 &
\underline{14.95 $\pm$ 0.33} &
25.01 $\pm$ 0.28 &
26.01 $\pm$ $1.94^\dag$  & 
\textbf{9.09 $\pm$ 0.45} \\
& $\rho \uparrow$ &
0.89 $\pm$ 0.01 &
0.91 $\pm$ $0.00^\dag$ &
0.94 $\pm$ 0.00 &
0.78 $\pm$ $0.00^\dag$ &
0.71 $\pm$ $0.02^\dag$ &
0.93 $\pm$ 0.01 &
\underline{0.95 $\pm$ 0.00} &
\underline{0.95 $\pm$ 0.00} &
0.94 $\pm$ $0.01^\dag$ & 
\textbf{0.98 $\pm$ 0.00} \\
& $\tau \uparrow$ &
0.74 $\pm$ 0.01 &
0.76 $\pm$ $0.00^\dag$ &
0.80 $\pm$ 0.01 &
0.61 $\pm$ $0.00^\dag$ &
0.54 $\pm$ $0.02^\dag$ &
0.78 $\pm$ 0.01 &
\underline{0.84 $\pm$ 0.00} &
0.82 $\pm$ 0.00 &
0.80 $\pm$ $0.02^\dag$ &
\textbf{0.89 $\pm$ 0.00} \\
& $p@10 \uparrow$ &
0.40 $\pm$ 0.05 &
0.46 $\pm$ $0.01^\dag$ &
0.71 $\pm$ 0.01 &
0.13 $\pm$ $0.01^\dag$ &
0.33 $\pm$ $0.05^\dag$ &
0.62 $\pm$ 0.05 &
0.71 $\pm$ 0.01 &
\underline{0.76 $\pm$ 0.00} &
0.69 $\pm$ $0.03^\dag$ &
\textbf{0.79 $\pm$ 0.01} \\
\midrule
\multirow{5}{*}{\textsc{PTC-FR}} & 
MSE $\downarrow$ &
1.44 $\pm$ 0.05 &
1.02 $\pm$ $0.02^\dag$ &
1.27 $\pm$ 0.05 &
4.97 $\pm$ 0.07 &
5.35 $\pm$ $0.35^\dag$ &
1.03 $\pm$ 0.09 &
\underline{0.40 $\pm$ 0.03} &
1.01 $\pm$ 0.04 &
1.11 $\pm$ $0.13^\dag$ &
\textbf{0.29 $\pm$ 0.01} \\
& MAE $\downarrow$ &
28.22 $\pm$ 0.73 &
18.07 $\pm$ $0.73^\dag$ &
25.86 $\pm$ 0.77 &
56.92 $\pm$ 2.04 &
56.92 $\pm$ $2.04^\dag$ &
23.47 $\pm$ 1.26 &
\underline{10.83 $\pm$ 0.70} &
22.95 $\pm$ 0.23 &
24.49 $\pm$ $1.07^\dag$ &
\textbf{8.69 $\pm$ 0.59} \\
& $\rho \uparrow$ &
0.89 $\pm$ 0.00 &
0.92 $\pm$ $0.00^\dag$ &
0.92 $\pm$ 0.00 &
0.75 $\pm$ 0.00 &
0.71 $\pm$ $0.02^\dag$ &
0.92 $\pm$ 0.01 &
\underline{0.96 $\pm$ 0.00} &
0.94 $\pm$ 0.00 &
0.94 $\pm$ $0.01^\dag$ &
\textbf{0.97 $\pm$ 0.00} \\
& $\tau \uparrow$ &
0.73 $\pm$ 0.00 &
0.79 $\pm$ $0.00^\dag$ &
0.78 $\pm$ 0.01 &
0.58 $\pm$ 0.00 &
0.53 $\pm$ $0.02^\dag$ &
0.77 $\pm$ 0.01 &
\underline{0.86 $\pm$ 0.01} &
0.81 $\pm$ 0.00 &
0.80 $\pm$ $0.02^\dag$ &
\textbf{0.88 $\pm$ 0.00} \\
& $p@10 \uparrow$ &
0.50 $\pm$ 0.01 &       
0.54 $\pm$ $0.03^\dag$ &   
0.72 $\pm$ 0.01 &   
0.16 $\pm$ 0.00 &
0.40 $\pm$ $0.02^\dag$ &  
0.65 $\pm$ 0.04 &    
\underline{0.78 $\pm$ 0.01} &     
0.77 $\pm$ 0.01 &    
0.69 $\pm$ $0.03^\dag$ &   
\textbf{0.81 $\pm$ 0.00} \\
\bottomrule
\end{tabular}}
\end{table*}

\subsubsection{Baselines} We compared our model against representative neural-based graph similarity learning models of two types : (i) models that rely on pairwise node/edge comparisons: SimGNN \cite{Bai2019SimGNNAN}, GraphSim \cite{graphsim}, GMN \cite{Li2019GraphMN}, GOTSim \cite{Doan2021InterpretableGS}, LMCCS \cite{Roy2022MaximumCS}, XMCS \cite{Roy2022MaximumCS}, ERIC \cite{zhuo2022efficient}; (ii) models solely based on graph embedding, including GEN \cite{Li2019GraphMN} and GREED \cite{ranjan2022greed}. We omitted the newest GEDGNN \cite{vldb} in our evaluation because it utilizes the ground truth matching matrix as a supervision signal, which is not the case for the other models. We used the official implementation and hyperparameters provided by the authors. For baselines \cite{Li2019GraphMN, Roy2022MaximumCS, ranjan2022greed} with similar but different training objectives, we carefully normalized their predicted score to ensure fair competition. 

\subsubsection{Implementation Details} We utilized Graph Isomorphism Networks (GIN) \cite{Xu2018HowPA} with skip connections as our Node-to-Region encoder, and ReLU as the activation. The encoder comprised 8 layers whose output dimension is 64. For \emph{Multi-sink Propagation}, we computed 5 flow paths of length 3 for each node (of length 6 for AIDS700). A two-layer MLP then computed 64-dimensional relative positions. We summarized the graph region using sum pooling and applied a linear layer to reduce its dimension to 32. We further transformed the concatenated shape of the regions into a scalar score using a two-layer $MLP(\cdot)_{t\in{MCS,GED}}$. We optimized the model using Adam with a fixed learning rate of 0.001. 

\subsubsection{Experimental Protocol} 
All models were trained on a single GeForce RTX 3090 GPU, hosted on a server with an Intel(R) Xeon(R) Silver 4210 CPU. All models are implemented in PyTorch. We utilized Python 3.9.16, PyTorch 1.12.1, PyTorch Geometric 2.2.0, and operated within the Ubuntu 20.04.6 LTS environment. We partitioned the datasets into training and test sets at a ratio of 4:1, with 20\% of the training set serving as the validation set. The batch size is set to 128, and each epoch consists of 100 iterations due to the abundance of graph pairs. Each model is trained for 50 epochs as a warm-up phase and then tested on the validation set every 20 epochs. We use early stopping with a patience of 50 to prevent overfitting, which means the validation loss fails to decrease for 50 consecutive validation steps. We set a maximum duration of 8 hours for each run.

\subsection{Results} 

\subsubsection{Results on MCS similarity learning}
Table \ref{tab:mcs} presents the results for MCS similarity learning and ranking, demonstrating that \textsc{G2R} consistently outperforms baselines across evaluation metrics.  For score prediction, \textsc{G2R} achieves a median MSE of $0.29\cdot 10^{-3}$, $55\%$ ($0.36\cdot 10^{-3}$ absolute) lower than baselines, and a median $p@10$ of $81\%$ in ranking, $10\%$ higher than others. The only exception is the ranking metric on the \emph{IMDB-Binary} dataset, which we further analyze in the ablation studies. ERIC, which constrains node-graph alignment during training, ranks as the second-best model. GraphSim that preserves structural information, however, is sensitive to the seed used. The seed can affect its initialization and node ordering scheme and leads to unstable performance. The results from GMN and GEN do not definitively demonstrate that fine-grained comparison improves performance over coarse-grained estimation. In fact, fine-grained comparison can introduce noise, as evidenced by the superior ranking performance of GEN compared to GMN. Finally, GOTsim performs poorly due to its reliance on a CPU-based combinatorial solver, which lacks parallelization and severely slows down training. This bottleneck prevents the model from converging within the designated 8-hour time frame.

\begin{figure*}[t]
\centering
\begin{minipage}{\linewidth}
\begin{minipage}{0.21\linewidth}
\centering
\setcounter{subfigure}{0}
\begin{subfigure}{1.0\linewidth}
\includegraphics[width=\linewidth]{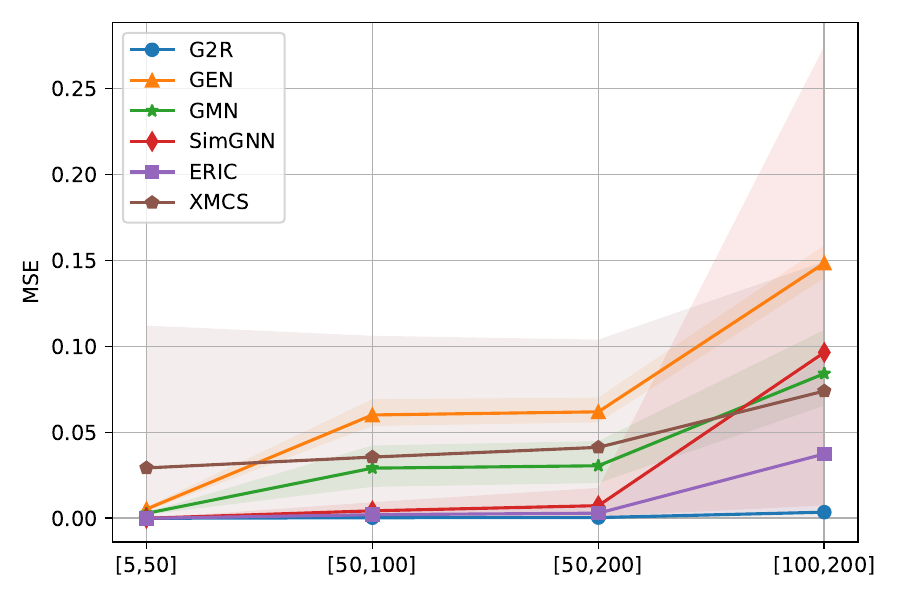}
\end{subfigure}\\
\begin{subfigure}{1.0\linewidth}
\includegraphics[width=\linewidth]{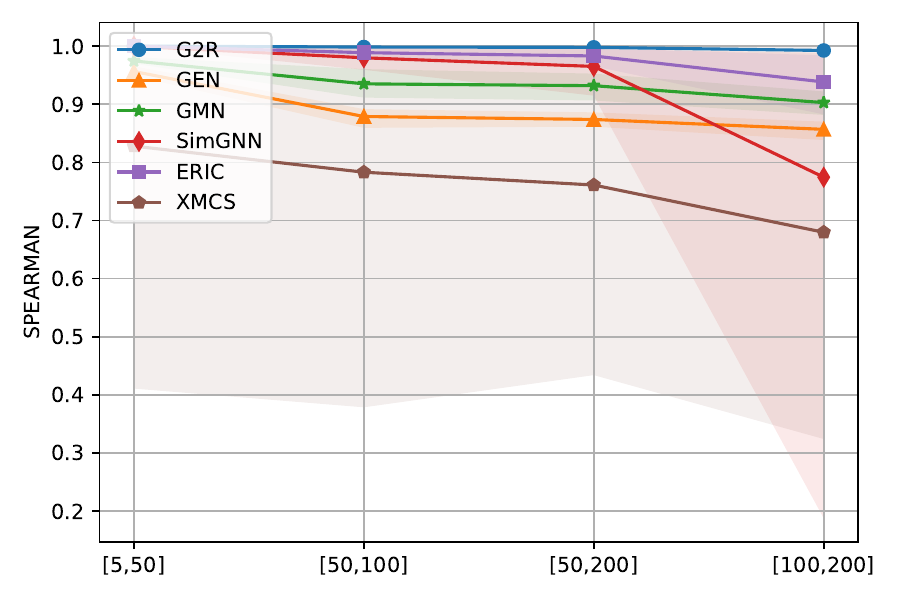}
\caption{\footnotesize MSRC\_21}
\end{subfigure}
\end{minipage}
\begin{minipage}{0.21\linewidth}
\centering
\begin{subfigure}{1.0\linewidth}
\includegraphics[width=\linewidth]{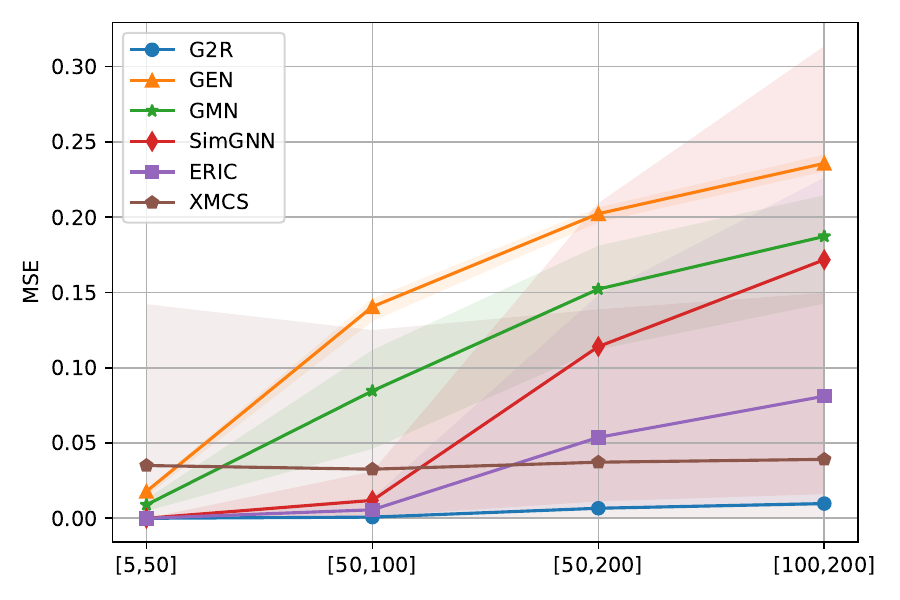}
\end{subfigure}
\begin{subfigure}{1.0\linewidth}
\includegraphics[width=\linewidth]{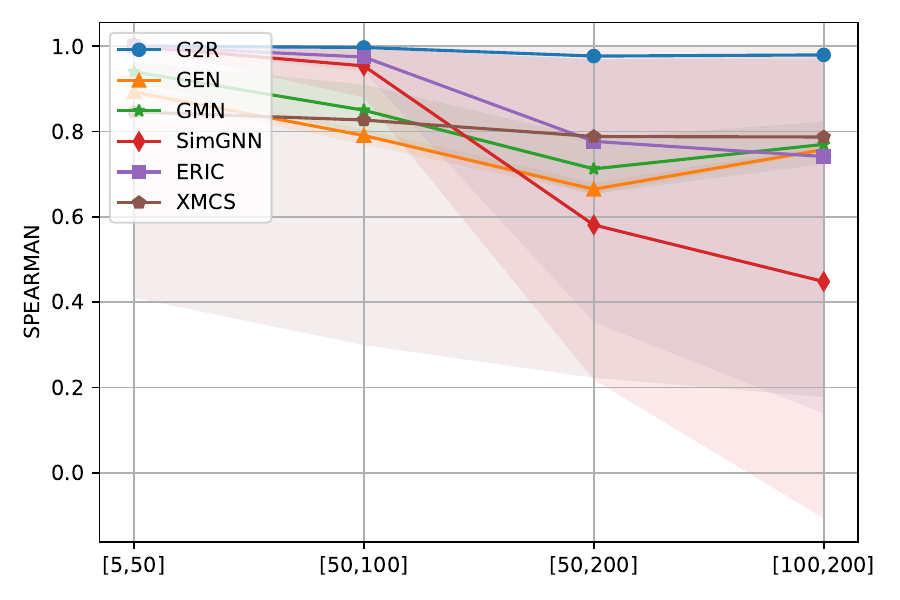}
\caption{\footnotesize D\&D}
\end{subfigure}
\end{minipage}
\begin{minipage}{0.21\linewidth}
\centering
\begin{subfigure}{1.0\linewidth}
\includegraphics[width=\linewidth]{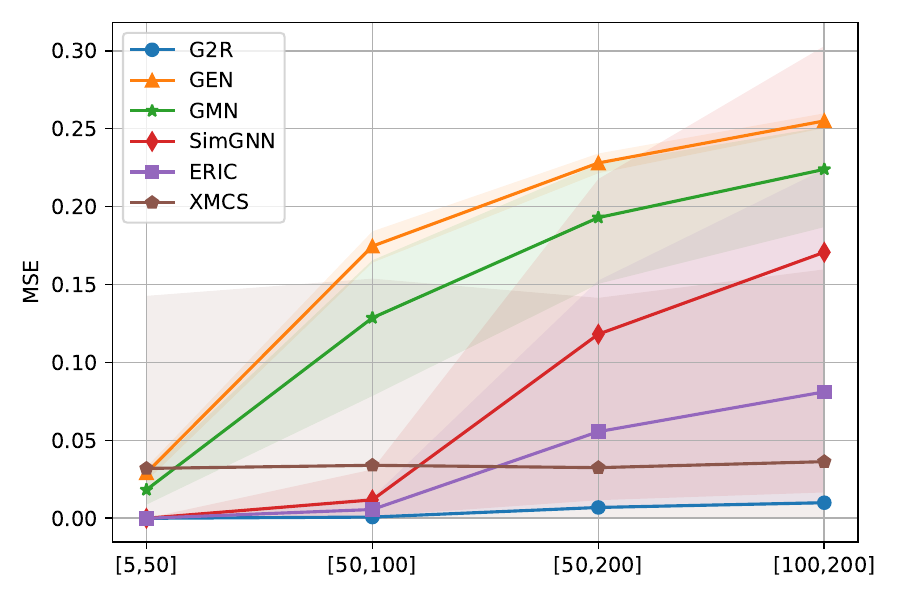}
\end{subfigure}
\begin{subfigure}{1.0\linewidth}
\includegraphics[width=\linewidth]{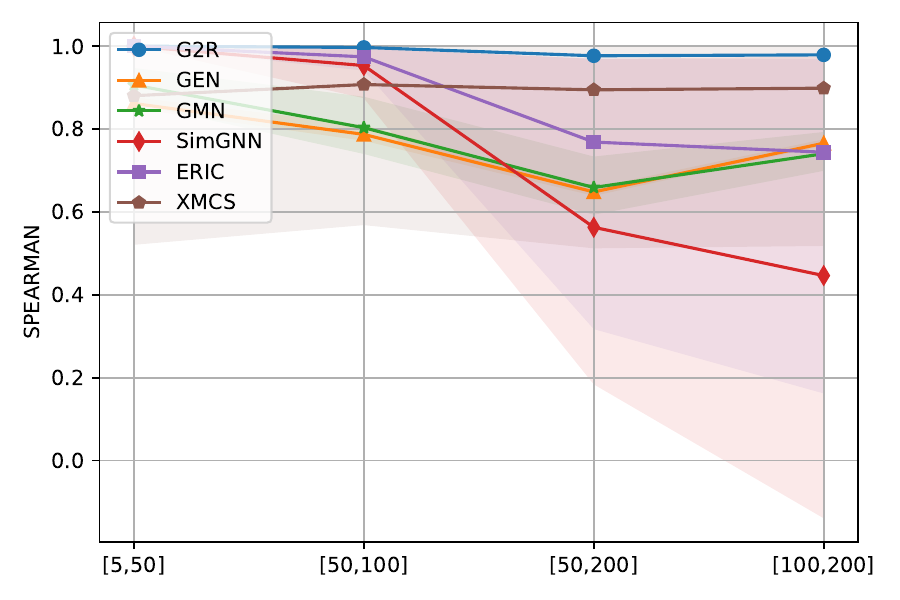}
\caption{\footnotesize FirstMM\_DB}
\end{subfigure}
\end{minipage}
\begin{minipage}{0.35\linewidth}
\centering
\begin{subfigure}{1.0\linewidth}
\includegraphics[width=\linewidth]{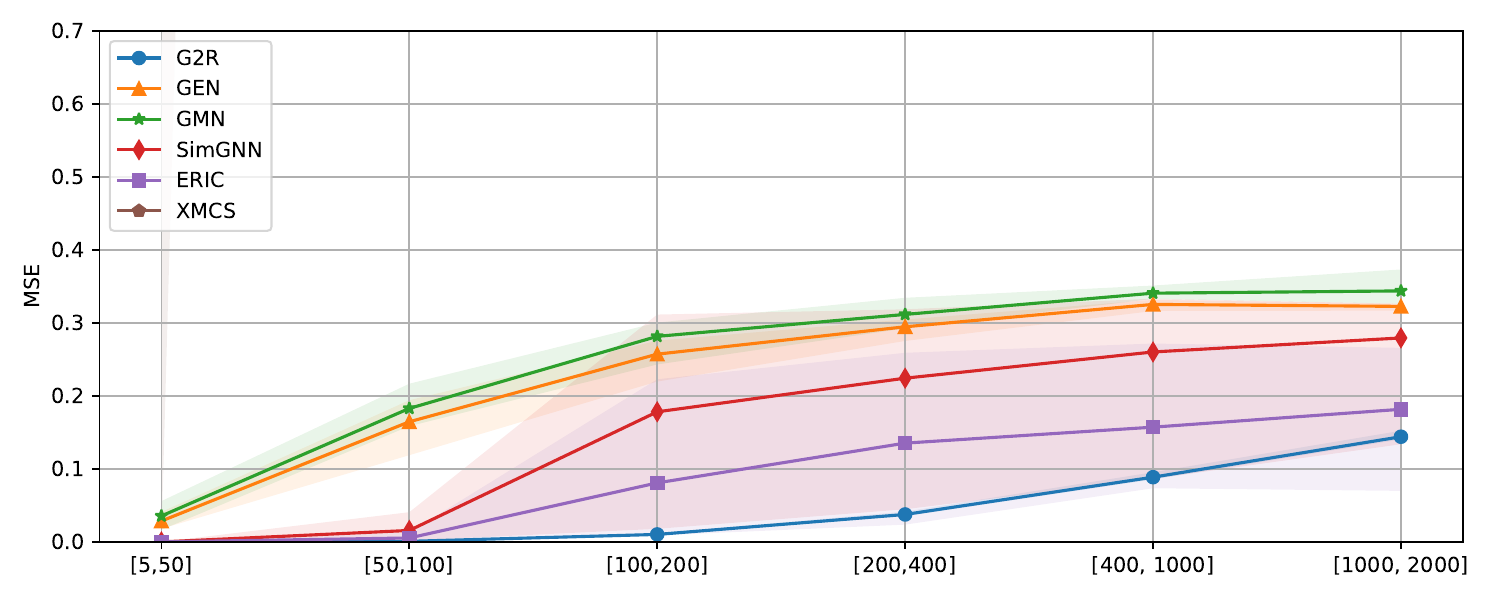}
\end{subfigure}
\begin{subfigure}{1.0\linewidth}
\includegraphics[width=\linewidth]{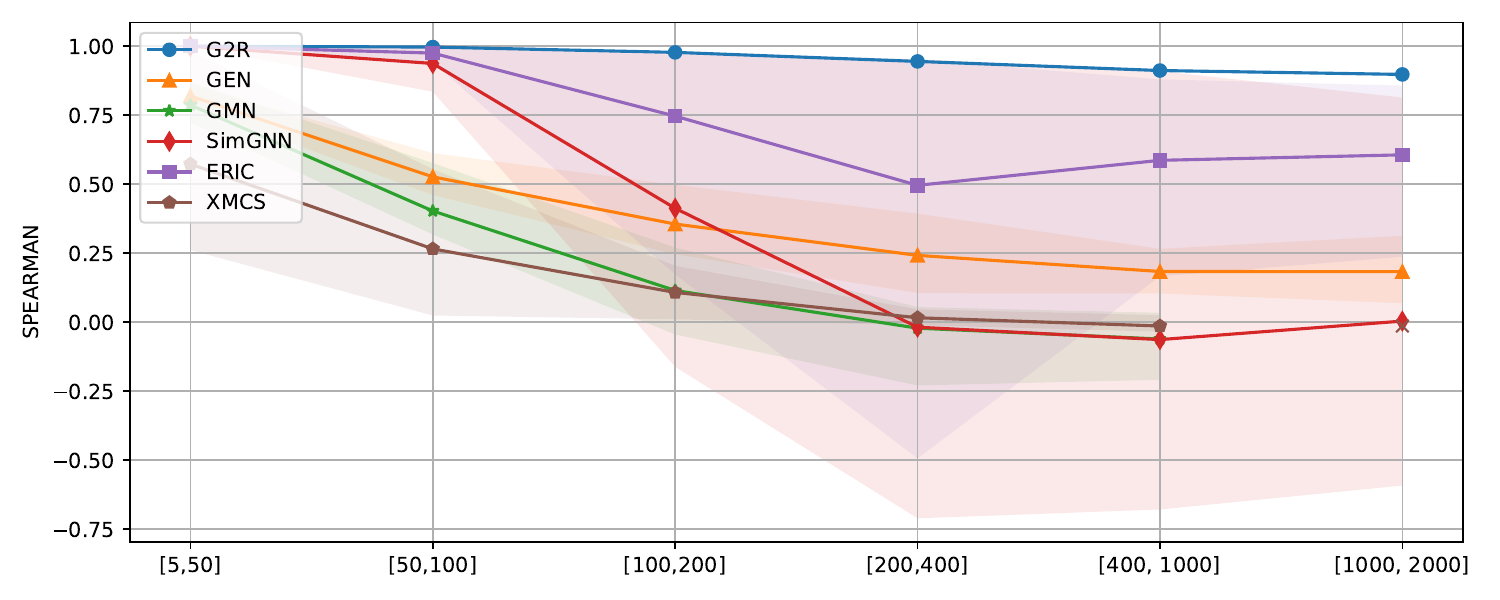}
\caption{\footnotesize WordNet}
\end{subfigure}
\end{minipage}
\caption{MSE (Up) and Spearman $\rho$ (Bottom) for each model across three real-world datasets.} \label{fig:trans}
\end{minipage}
\end{figure*}

\begin{table*}[ht]
\caption{Prediction and ranking of MCS similarity on test sets. The MSE and MAE are in $10^{-3}$. The best is highlighted in bold, while the second is underlined. Results marked with $\dag$ mean we report the best five runs among at least ten.}\label{tab:results_mcs}
\resizebox{\textwidth}{!}{ \renewcommand{\arraystretch}{1.0}
\centering
\begin{tabular}{@{}cl|ccccc|ccccc|ccccc}
\cmidrule[\heavyrulewidth]{2-17}

\multirow{2}{*}{} &
\multirow{2}{*}{\textbf{Dataset}} &
\multicolumn{5}{c}{\textsc{AIDS700}} &
\multicolumn{5}{c}{\textsc{Linux}} &
\multicolumn{5}{c}{\textsc{IMDB-Multi}}
\\
\cmidrule(l{2pt}r{2pt}){3-17}
& & 
MSE$\downarrow$ & MAE$\downarrow$ & $\rho$$\uparrow$ & $\tau$$\uparrow$ & 
p@10$\uparrow$ &
MSE$\downarrow$ & MAE$\downarrow$ & $\rho$$\uparrow$ & $\tau$$\uparrow$ &
p@10$\uparrow$ &
MSE$\downarrow$ & MAE$\downarrow$ & $\rho$$\uparrow$ & $\tau$$\uparrow$ &
p@10$\uparrow$ \\
\cmidrule(l{2pt}r{2pt}){2-17}

\multirow{9}{*}{\rotatebox{90}{Baselines}} 
& \textsc{SimGNN} \cite{Bai2019SimGNNAN} &
3.55 $\pm$ 0.20 &
46.24 $\pm$ 1.28 &
0.61 $\pm$ 0.02 &
0.45 $\pm$ 0.01 &
0.28 $\pm$ 0.03 &

0.91 $\pm$ 0.24 &
19.61 $\pm$ 3.23 &
0.90 $\pm$ 0.02 &
0.74 $\pm$ 0.03 &
0.81 $\pm$ 0.06 &

0.63 $\pm$ 0.12 &
16.14 $\pm$ 19.84 &
0.97 $\pm$ 0.00 &
0.88 $\pm$ 0.01 &
0.82 $\pm$ 0.02 \\

& \textsc{GraphSim} \cite{graphsim} &
3.42 $\pm$ $0.15^\dag$ &
44.59 $\pm$ $1.00^\dag$ &
0.63 $\pm$ $0.01^\dag$ &
0.47 $\pm$ $0.01^\dag$ &
0.37 $\pm$ $0.03^\dag$ &

0.21 $\pm$ $0.02^\dag$ &
4.40 $\pm$ $0.64^\dag$  &
\textbf{0.97 $\pm$} $\textbf{0.00}^\dag$  &
\textbf{0.86 $\pm$} $\textbf{0.00}^\dag$  &
0.95 $\pm$ $0.01^\dag$  &

0.72 $\pm$  $0.04^\dag$ &
11.77 $\pm$ $0.60^\dag$ &
0.97 $\pm$ $0.00^\dag$ &
0.89 $\pm$ $0.00^\dag$ &
0.82 $\pm$ $0.00^\dag$ \\

& \textsc{GMN} \cite{Li2019GraphMN} &
2.03 $\pm$ 0.12 &
33.18 $\pm$ 0.71 &
0.82 $\pm$ 0.01 &
0.64 $\pm$ 0.01 &
0.73 $\pm$ 0.01 &

0.22 $\pm$ 0.04 &
8.36 $\pm$ 1.17 &
\textbf{0.97 $\pm$ 0.00} &
\underline{0.84 $\pm$ 0.01} &
0.95 $\pm$ 0.01 & 

0.21 $\pm$ 0.03 &
8.73 $\pm$ 0.83 &
0.98 $\pm$ 0.00 &
\underline{0.91 $\pm$ 0.00} &
\underline{0.91 $\pm$ 0.01} \\

& \textsc{GOTSim} \cite{Doan2021InterpretableGS} &
6.10 $\pm$ 0.11 &
61.86 $\pm$ 0.43 &
0.32 $\pm$ 0.00 &
0.23 $\pm$ 0.00 &
0.08 $\pm$ 0.00 &

5.27 $\pm$ 0.05 &
58.67 $\pm$ 0.32 &
0.52 $\pm$ 0.00 &
0.39 $\pm$ 0.01 &
0.24 $\pm$ 0.04 &

13.40 $\pm$ 0.04 &
91.47 $\pm$ 1.52 &
0.58 $\pm$ 0.00 &
0.46 $\pm$ 0.00 &
0.49 $\pm$ 0.01\\

& \textsc{LMCCS} \cite{Roy2022MaximumCS} &
6.17 $\pm$ 1.33 &
59.19 $\pm$ 5.23 &
0.38 $\pm$ 0.09 &
0.27 $\pm$ 0.07 &
0.19 $\pm$ 0.15 &

1.95 $\pm$ 0.18 &
33.12 $\pm$ 1.66 &
0.80 $\pm$ 0.02 &
0.62 $\pm$ 0.02 &
0.89 $\pm$ 0.01 &

3.72 $\pm$ 0.38 &
45.64 $\pm$ 3.59 &
0.87 $\pm$ 0.01 &
0.71 $\pm$ 0.01 &
0.39 $\pm$ 0.02 \\

& \textsc{XMCS} \cite{Roy2022MaximumCS} &
2.17 $\pm$ 0.04 &
36.70 $\pm$ 0.40 &
0.72 $\pm$ 0.01 &
0.54 $\pm$ 0.05 &
0.50 $\pm$ 0.02 &

0.82 $\pm$ 0.21 &
19.58 $\pm$ 2.72 &
0.91 $\pm$ 0.02 &
0.74 $\pm$ 0.03 &
0.88 $\pm$ 0.02 &

1.40 $\pm$ $0.31^\dag$ &
26.90 $\pm$ $3.51^\dag$ &
0.94 $\pm$ $0.01^\dag$ &
0.81 $\pm$ $0.02^\dag$ &
0.63 $\pm$ $0.06^\dag$ \\

& \textsc{ERIC} \cite{zhuo2022efficient} &
2.94 $\pm$ 0.30 &
41.62 $\pm$ 2.00 &
0.66 $\pm$ 0.03 &
0.49 $\pm$ 0.02 &
0.37 $\pm$ 0.06 &

0.10 $\pm$ 0.01 &
3.10 $\pm$ 0.26 &
\textbf{0.97 $\pm$ 0.00} &
\textbf{0.86 $\pm$ 0.00} &
\textbf{0.98 $\pm$ 0.01} &

0.42 $\pm$ 0.03 &
10.64 $\pm$ 0.77 &
\underline{0.98 $\pm$ 0.00} &
0.90 $\pm$ 0.00 &
0.86 $\pm$ 0.01\\

\cmidrule{2-17}

& \textsc{GEN} \cite{Li2019GraphMN} &
1.77 $\pm$ 0.07 &
30.33 $\pm$ 0.23 &
0.84 $\pm$ 0.00 &
0.66 $\pm$ 0.00 &
\underline{0.74 $\pm$ 0.01} &

0.35 $\pm$ 0.02 &
12.57 $\pm$ 0.36 &
\underline{0.95 $\pm$ 0.00} &
0.81 $\pm$ 0.01 &
0.95 $\pm$ 0.00 &

0.26 $\pm$ 0.02 &
9.36 $\pm$ 0.68 &
\underline{0.98 $\pm$ 0.00} &
\underline{0.91 $\pm$ 0.00} &
0.89 $\pm$ 0.00 \\

& \textsc{Greed}  \cite{ranjan2022greed} &
2.20 $\pm$ $0.22^\dag$ &
36.71 $\pm$ $1.55^\dag$ &
0.76 $\pm$ $0.01^\dag$ &
0.58 $\pm$ $0.01^\dag$ &
0.55 $\pm$ $0.03^\dag$ &

5.83 $\pm$ 2.71 &
57.11 $\pm$ 21.73 &
0.23 $\pm$ 0.36 &
0.19 $\pm$ 0.31 &
0.24 $\pm$ 0.35 &

0.75 $\pm$ $0.23^\dag$ &
17.61 $\pm$ $3.56^\dag$ &
0.97 $\pm$ $0.01^\dag$ &
0.88 $\pm$ $0.01^\dag$ &
0.85 $\pm$ $0.01^\dag$ \\

\cmidrule{2-17}

\multirow{5}{*}{\rotatebox{90}{Ablation} } 
& \textsc{w/o PE} &
1.99 $\pm$ 0.45 &
27.04 $\pm$ 1.95 &
0.84 $\pm$ 0.01 &
0.68 $\pm$ 0.02 &
0.68 $\pm$ 0.03 &

0.20 $\pm$ 0.06 &
3.31 $\pm$ 0.51 &
\textbf{0.97 $\pm$ 0.00} &
\textbf{0.86 $\pm$ 0.00} &
\underline{0.97 $\pm$ 0.00} &

0.15 $\pm$ 0.07 &
6.03 $\pm$ 0.24 &
\textbf{0.99 $\pm$ 0.00} &
\textbf{0.92 $\pm$ 0.00} &
\textbf{0.92 $\pm$ 0.00} \\

& \textsc{w/o Shape Score} &
1.49 $\pm$ 0.08 &
30.01 $\pm$ 0.73 &
0.82 $\pm$ 0.01 &
0.65 $\pm$ 0.01 &
0.73 $\pm$ 0.02 &

0.46 $\pm$ 0.02 &
15.75 $\pm$ 0.26 &
0.94 $\pm$ 0.00 &
0.78 $\pm$ 0.00 &
0.95 $\pm$ 0.00 &

0.39 $\pm$ 0.01 &
11.64 $\pm$ 0.21 &
0.97 $\pm$ 0.00 &
0.88 $\pm$ 0.00 &
0.88 $\pm$ 0.01 \\

& \textsc{w/o Vol Score} & 
2.20 $\pm$ 0.51 &
29.03 $\pm$ 6.23 &
0.81 $\pm$ 0.03 &
0.65 $\pm$ 0.04 &
0.60 $\pm$ 0.08 &

0.15 $\pm$ 0.01 &
2.94 $\pm$ 0.37 &
\textbf{0.97 $\pm$ 0.00} &
\textbf{0.86 $\pm$ 0.00} &
\underline{0.97 $\pm$ 0.00} &

0.27 $\pm$ 0.04 &
8.13 $\pm$ 1.00 &
\underline{0.98 $\pm$ 0.00} &
0.90 $\pm$ 0.00 &
0.87 $\pm$ 0.02 \\

& \textsc{No Clamp} &
2.23 $\pm$ 0.53 &
26.04 $\pm$ 2.53 &
0.85 $\pm$ 0.02 &
0.69 $\pm$ 0.02 &
0.72 $\pm$ 0.04 &

0.20 $\pm$ 0.04 &
3.89 $\pm$ 1.07 &
\textbf{0.97 $\pm$ 0.00} &
\textbf{0.86 $\pm$ 0.00} &
0.96 $\pm$ 0.01 &

0.17 $\pm$ 0.02 &
6.39 $\pm$ 0.23 &
\underline{0.98 $\pm$ 0.00} &
0.90 $\pm$ 0.00 &
\underline{0.91 $\pm$ 0.01} \\

& \textsc{w/o Uncertainty} &
1.20 $\pm$ 0.03 &
24.11 $\pm$ 0.51 &
0.85 $\pm$ 0.00 &
0.69 $\pm$ 0.01 &
0.73 $\pm$ 0.01 &

\underline{0.09 $\pm$ 0.02} &
\textbf{2.20 $\pm$ 0.27} &
\textbf{0.97 $\pm$ 0.00} &
\textbf{0.86 $\pm$ 0.00} &
\underline{0.97 $\pm$ 0.01} &

0.16 $\pm$ 0.02 &
6.02 $\pm$ 0.42 &
\underline{0.98 $\pm$ 0.00} &
0.90 $\pm$ 0.00 &
0.90 $\pm$ 0.01 \\

\cmidrule{2-17}
& \textsc{G2R} &
\textbf{0.84 $\pm$ 0.03} &
\textbf{18.88 $\pm$ 0.68} &
\textbf{0.89 $\pm$ 0.01} &
\textbf{0.73 $\pm$ 0.01} &
\textbf{0.81 $\pm$ 0.01} &

\textbf{0.08 $\pm$ 0.01} &
\underline{2.30 $\pm$ 0.29} &
\textbf{0.97 $\pm$ 0.00} &
\textbf{0.86 $\pm$ 0.00} &
\underline{0.97 $\pm$ 0.01} &

\underline{0.14 $\pm$ 0.03} &
\textbf{5.26 $\pm$ 0.61} &
\underline{0.98 $\pm$ 0.00} &
0.90 $\pm$ 0.00 &
\textbf{0.92 $\pm$ 0.01} \\

& \textsc{G2R-Dual} &
\underline{1.06 $\pm$ 0.04} &
\underline{22.59 $\pm$ 0.82} &
\underline{0.86 $\pm$ 0.01} &
\underline{0.70 $\pm$ 0.01} &
\underline{0.74 $\pm$ 0.01} &

\underline{0.09 $\pm$ 0.01} &
2.46 $\pm$ 0.28 &
\textbf{0.97 $\pm$ 0.00} &
\textbf{0.86 $\pm$ 0.00} &
\underline{0.97 $\pm$ 0.00} &

0.17 $\pm$ 0.01 &
6.50 $\pm$ 0.35 &
\underline{0.98 $\pm$ 0.00} &
0.90 $\pm$ 0.00 &
\underline{0.91 $\pm$ 0.01} \\

& \textsc{Shared MLP} &
1.12 $\pm$ 0.08 &
23.61 $\pm$ 0.95 &
\underline{0.86 $\pm$ 0.01} &
0.69 $\pm$ 0.01 &
0.75 $\pm$ 0.03 &

0.10 $\pm$ 0.02 &
2.78 $\pm$ 0.20 &
\textbf{0.97 $\pm$ 0.00} &
\textbf{0.86 $\pm$ 0.00} &
\underline{0.97 $\pm$ 0.01} &

\textbf{0.13 $\pm$ 0.02} &
\underline{5.29 $\pm$ 0.41} &
\underline{0.98 $\pm$ 0.00} &
0.90 $\pm$ 0.00 &
\underline{0.91 $\pm$ 0.01} \\
\cmidrule[\heavyrulewidth]{2-17}
\end{tabular}
}
\end{table*}

\begin{table*}[ht]
\caption{Prediction and ranking of GED similarity on test sets. The MSE and MAE are in $10^{-3}$. The best is highlighted in bold, while the second is underlined. Results marked with $\dag$ mean we report the best five runs among at least ten.}\label{tab:results_ged}
\resizebox{\textwidth}{!}{ \renewcommand{\arraystretch}{1.0}
\centering
\begin{tabular}{@{}cl|ccccc|ccccc|ccccc}
\cmidrule[\heavyrulewidth]{2-17}
\multirow{2}{*}{} &
\multirow{2}{*}{\textbf{Dataset}} &
\multicolumn{5}{c}{\textsc{AIDS700}} &
\multicolumn{5}{c}{\textsc{Linux}} &
\multicolumn{5}{c}{\textsc{IMDB-Multi}}
\\
\cmidrule(l{2pt}r{2pt}){3-17}
& & 
MSE$\downarrow$ & MAE$\downarrow$ & $\rho$$\uparrow$ & $\tau$$\uparrow$ & 
p@10$\uparrow$ &
MSE$\downarrow$ & MAE$\downarrow$ & $\rho$$\uparrow$ & $\tau$$\uparrow$ &
p@10$\uparrow$ &
MSE$\downarrow$ & MAE$\downarrow$ & $\rho$$\uparrow$ & $\tau$$\uparrow$ &
p@10$\uparrow$ \\
\cmidrule(l{2pt}r{2pt}){2-17}
\multirow{9}{*}{\rotatebox{90}{Baselines}} 
& \textsc{SimGNN} \cite{Bai2019SimGNNAN} &
1.97 $\pm$ 0.05 &
32.79 $\pm$ 0.43 &
0.87 $\pm$ 0.00 &
0.70 $\pm$ 0.00 &
0.50 $\pm$ 0.01 &

2.00 $\pm$ 0.36 &
31.91 $\pm$ 3.86 &
0.95 $\pm$ 0.01 &
0.80 $\pm$ 0.01 &
0.89 $\pm$ 0.04 &

1.69 $\pm$ 0.19 &
19.84 $\pm$ 1.73 &
0.86 $\pm$ 0.06 &
0.73 $\pm$ 0.06 &
0.78 $\pm$ 0.01\\

& \textsc{GraphSim} \cite{graphsim} &
2.15 $\pm$ $0.05^\dag$ &
34.54 $\pm$ $0.56^\dag$ &
0.86 $\pm$ $0.00^\dag$ &
0.69 $\pm$ $0.00^\dag$ &
0.48 $\pm$ $0.02^\dag$ &

0.20 $\pm$ $0.02^\dag$ & 
4.69 $\pm$ $0.53^\dag$ &
\textbf{0.99 $\pm$} $\textbf{0.00}^\dag$ &
\textbf{0.91 $\pm$} $\textbf{0.00}^\dag$ &
\textbf{0.99 $\pm$} $\textbf{0.00}^\dag$ &

1.44 $\pm$ $0.13^\dag$ &
16.51 $\pm$ $0.89^\dag$ &
0.85 $\pm$ $0.01^\dag$ &
0.72 $\pm$ $0.01^\dag$ &
0.78 $\pm$ $0.00^\dag$ \\

& \textsc{GMN} \cite{Li2019GraphMN} &
4.26 $\pm$ 0.17 &
48.37 $\pm$ 0.68 &
0.85 $\pm$ 0.01 &
0.68 $\pm$ 0.01 &
0.59 $\pm$ 0.01 &

0.48 $\pm$ 0.10 &
12.97 $\pm$ 1.56 &
\underline{0.98 $\pm$ 0.00} &
0.88 $\pm$ 0.01 &
0.97 $\pm$ 0.01 &

0.48 $\pm$ 0.04 &
9.37 $\pm$ 0.86 &
\underline{0.92 $\pm$ 0.01} &
\textbf{0.81 $\pm$ 0.02} &
0.87 $\pm$ 0.00 \\

& \textsc{GOTSim} \cite{Doan2021InterpretableGS} &
8.88 $\pm$ 1.30 &
72.93 $\pm$ 4.70 &
0.49 $\pm$ 0.10 &
0.36 $\pm$ 0.08 &
0.10 $\pm$ 0.05 &

10.28 $\pm$ 0.35 &
76.59 $\pm$ 1.80 &
0.86 $\pm$ 0.00 &
0.68 $\pm$ 0.00 &
0.25 $\pm$ 0.01 &

20.26 $\pm$ $0.22^\dag$ &
89.91 $\pm$ $0.46^\dag$ &
0.78 $\pm$ $0.00^\dag$ &
0.62 $\pm$ $0.00^\dag$ &
0.52 $\pm$ $0.01^\dag$ \\

& \textsc{LMCCS} \cite{Roy2022MaximumCS} &
7.97 $\pm$ 1.39 &
67.59 $\pm$ 5.84 &
0.48 $\pm$ 0.07 &
0.35 $\pm$ 0.05 &
0.07 $\pm$ 0.05 &

4.45 $\pm$ $0.30^\dag$ &
44.83 $\pm$ $2.17^\dag$ &
0.90 $\pm$ $0.00^\dag$ &
0.74 $\pm$ $0.01^\dag$ &
0.89 $\pm$ $0.02^\dag$ &

6.02 $\pm$ 0.71 &
39.79 $\pm$ 2.06 &
0.73 $\pm$ 0.02 &
0.60 $\pm$ 0.02 &
0.58 $\pm$ 0.05\\

& \textsc{XMCS} \cite{Roy2022MaximumCS} &
1.97 $\pm$ 0.21 &
33.38 $\pm$ 1.52 &
0.86 $\pm$ 0.01 &
0.69 $\pm$ 0.01 &
0.52 $\pm$ 0.05 &

1.10 $\pm$ $0.27^\dag$  &
24.13 $\pm$ $3.08^\dag$  &
0.96 $\pm$ $0.01^\dag$  &
0.83 $\pm$ $0.01^\dag$  &
0.96 $\pm$ $0.01^\dag$  &

1.31 $\pm$ $0.22^\dag$  &
18.03 $\pm$ $1.74^\dag$  &
0.91 $\pm$ $0.01^\dag$  &
0.78 $\pm$ $0.02^\dag$  &
0.83 $\pm$ $0.02^\dag$  \\

& \textsc{ERIC} \cite{zhuo2022efficient} &
1.68 $\pm$ 0.03 &
30.67 $\pm$ 0.20 &
0.88 $\pm$ 0.00 &
0.72 $\pm$ 0.00 &
0.57 $\pm$ 0.01 &

0.11 $\pm$ 0.02 &
2.96 $\pm$ 0.33 &
\textbf{0.99 $\pm$ 0.00} &
\textbf{0.91 $\pm$ 0.00} &
\textbf{0.99 $\pm$ 0.00} &

0.53 $\pm$ 0.09 &
9.47 $\pm$ 0.89 &
0.90 $\pm$ 0.01 &
\underline{0.79 $\pm$ 0.01} &
0.87 $\pm$ 0.01\\

& \textsc{GEN} \cite{Li2019GraphMN} &
4.29 $\pm$ 0.24 &
47.67 $\pm$ 1.59 &
0.87 $\pm$ 0.00 &
0.70 $\pm$ 0.01 &
0.60 $\pm$ 0.01 &

0.42 $\pm$ 0.01 &
10.68 $\pm$ 0.58 &
\underline{0.98 $\pm$ 0.00} &
0.88 $\pm$ 0.00 &
0.97 $\pm$ 0.00 &

0.96 $\pm$ $0.24^\dag$ &
14.46 $\pm$ $2.29^\dag$ &
0.87 $\pm$ $0.03^\dag$ &
0.75 $\pm$ $0.02^\dag$ &
0.85 $\pm$ $0.01^\dag$ \\

& \textsc{Greed}  \cite{ranjan2022greed} &
1.66 $\pm$ 0.03 &
30.71 $\pm$ 0.23 &
0.89 $\pm$ 0.00 &
0.73 $\pm$ 0.00 &
0.59 $\pm$ 0.01 &

0.87 $\pm$ 0.02 &
22.44 $\pm$ 0.12 &
0.97 $\pm$ 0.00 &
0.84 $\pm$ 0.00 &
0.97 $\pm$ 0.00 &

0.73 $\pm$ 0.01 &
13.40 $\pm$ 0.09 &
0.91 $\pm$ 0.01 &
\underline{0.79 $\pm$ 0.01} &
0.85 $\pm$ 0.01 \\

\cmidrule{2-17}

\multirow{5}{*}{\rotatebox{90}{Ablation} } 
& \textsc{w/o PE} &
2.82 $\pm$ 0.07 &
39.14 $\pm$ 0.64 &
0.81 $\pm$ 0.00 &
0.65 $\pm$ 0.00 &
0.50 $\pm$ 0.01 &

1.05 $\pm$ 0.10 &
7.97 $\pm$ 1.07 &
0.97 $\pm$ 0.00 &
0.88 $\pm$ 0.00 &
0.97 $\pm$ 0.00 &

0.44 $\pm$ 0.01 &
9.44 $\pm$ 0.25 &
\textbf{0.93 $\pm$ 0.00} &
\textbf{0.81 $\pm$ 0.01} &
\underline{0.88 $\pm$ 0.00} \\

& \textsc{w/o Shape Score} &
1.71 $\pm$ 0.07 &
31.01 $\pm$ 0.47 &
0.89 $\pm$ 0.00 &
0.73 $\pm$ 0.00 &
0.61 $\pm$ 0.01 &

0.43 $\pm$ 0.01 &
13.61 $\pm$ 0.17 &
\underline{0.98 $\pm$ 0.00} &
0.87 $\pm$ 0.00 &
\underline{0.98 $\pm$ 0.00} &

0.54 $\pm$ 0.01 &
11.12 $\pm$ 0.11 &
0.90 $\pm$ 0.02 &
0.78 $\pm$ 0.02 &
0.86 $\pm$ 0.01 \\

& \textsc{w/o Vol Score} & 
1.47 $\pm$ 0.03 &
28.45 $\pm$ 0.22 &
\textbf{0.91 $\pm$ 0.00} &
\underline{0.75 $\pm$ 0.00} &
\underline{0.66 $\pm$ 0.01} &

\underline{0.07 $\pm$ 0.00} &
\underline{2.54 $\pm$ 0.09} &
\textbf{0.99 $\pm$ 0.00} &
\underline{0.90 $\pm$ 0.00} &
0.90 $\pm$ 0.00 &

0.46 $\pm$ 0.01 &
10.07 $\pm$ 0.44 &
0.90 $\pm$ 0.03 &
\underline{0.79 $\pm$ 0.02} &
0.87 $\pm$ 0.01 \\

& \textsc{No Clamp} &
3.13 $\pm$ 0.18 &
41.63 $\pm$ 0.81 &
0.80 $\pm$ 0.01 &
0.63 $\pm$ 0.00 &
0.48 $\pm$ 0.01 &

0.76 $\pm$ 0.33 &
6.82 $\pm$ 1.82 &
0.97 $\pm$ 0.01 &
0.89 $\pm$ 0.01 &
\underline{0.98 $\pm$ 0.01} &

\textbf{0.41 $\pm$ 0.02} &
\underline{8.70 $\pm$ 0.36} &
\underline{0.92 $\pm$ 0.01} &
\textbf{0.81 $\pm$ 0.01} &
\textbf{0.89 $\pm$ 0.01} \\

& \textsc{w/o Uncertainty} &
1.30 $\pm$ 0.06 &
26.99 $\pm$ 0.68 &
\textbf{0.91 $\pm$ 0.00} &
\underline{0.75 $\pm$ 0.00} &
\textbf{0.67 $\pm$ 0.01} &

0.09 $\pm$ 0.02 &
2.87 $\pm$ 0.63 &
\textbf{0.99 $\pm$ 0.00} &
\underline{0.90 $\pm$ 0.00} &
\textbf{0.99 $\pm$ 0.00} &

0.43 $\pm$ 0.02 &
8.93 $\pm$ 0.49 &
0.88 $\pm$ 0.01 &
0.77 $\pm$ 0.01 &
\textbf{0.89 $\pm$ 0.00}\\

\cmidrule{2-17}

& \textsc{G2R} &
1.30 $\pm$ 0.03 &
27.08 $\pm$ 0.43 &
\textbf{0.91 $\pm$ 0.00} &
\underline{0.75 $\pm$ 0.00} &
\underline{0.66 $\pm$ 0.01} &

\textbf{0.05 $\pm$ 0.01} &
\textbf{1.92 $\pm$ 0.39} &
\textbf{0.99 $\pm$ 0.00} &
\underline{0.90 $\pm$ 0.00} &
\textbf{0.99 $\pm$ 0.00} &

\underline{0.42 $\pm$ 0.01} &
\textbf{8.69 $\pm$ 0.14} &
\textbf{0.93 $\pm$ 0.01} &
\textbf{0.81 $\pm$ 0.01} &
\underline{0.88 $\pm$ 0.01} \\

& \textsc{G2R-Dual} &
\textbf{1.26 $\pm$ 0.02} &
\textbf{26.64 $\pm$ 0.21} &
\textbf{0.91 $\pm$ 0.00} &
\textbf{0.76 $\pm$ 0.00} &
\underline{0.66 $\pm$ 0.01} &

0.08 $\pm$ 0.01 &
2.59 $\pm$ 0.27 &
\textbf{0.99 $\pm$ 0.00} &
\underline{0.90 $\pm$ 0.00} &
\textbf{0.99 $\pm$ 0.00} &

0.43 $\pm$ 0.02 &
8.93 $\pm$ 0.49 &
0.88 $\pm$ 0.01&
0.77 $\pm$ 0.01 &
\textbf{0.89 $\pm$ 0.00}\\

& \textsc{Shared MLP} &
\underline{1.28 $\pm$ 0.03} &
\underline{26.79 $\pm$ 0.29} &
\textbf{0.91 $\pm$ 0.00} &
\underline{0.75 $\pm$ 0.00} &
\underline{0.66 $\pm$ 0.01} &

0.08 $\pm$ 0.01 &
2.64 $\pm$ 0.32 &
\textbf{0.99 $\pm$ 0.00} &
\underline{0.90 $\pm$ 0.00} &
\textbf{0.99 $\pm$ 0.00} &

\underline{0.42 $\pm$ 0.01} &
8.97 $\pm$ 0.22 &
0.89 $\pm$ 0.01 &
0.78 $\pm$ 0.01 &
\textbf{0.89 $\pm$ 0.01}\\

\cmidrule[\heavyrulewidth]{2-17}
\end{tabular}
}
\end{table*}

\subsubsection{Transferability} MCS and GED are commonly used to measure similarity in small graphs due to their NP-hard nature. To evaluate the transferability of models under extreme conditions, we train models using synthetic graphs within the [5,50] node size group, following \cite{vldb}, but increase the challenge by testing them on real-world datasets across different node size groups: [5,50], [50,100], [50,200], [100,200], [200,400], [400,1000], and [1000,2000]. We select baselines that demonstrated comparable and stable performance in MCS similarity learning. Figure \ref{fig:trans} illustrates the MSE and ranking results across these groups. For score prediction, \textsc{G2R} remains the top performer, although MSE increases, especially in groups with node sizes beyond [200, 400]. This suggests that, while \textsc{G2R} demonstrates good transferability, it requires either training on larger graphs than [5,50] or fine-tuning on larger graphs to maintain strong zero-shot prediction performance  on larger graphs. The Gumbel-Sinkhorn network enhances XMCS with the smallest increase in MSE on the \emph{D\&D} and \emph{FirstMM\_DB} datasets, but performs the worst on the \emph{WordNet} dataset. This can be attributed to the denser structure of \emph{WordNet}, which introduces different distributions that cause XMCS performance to drop significantly. For ranking prediction, \textsc{G2R} exhibits remarkable transferability, outperforming the baselines. This success is due to \textsc{G2R}'s ability to model graph scale. By incorporating volume scores that account for the overlap region of input pairs, \textsc{G2R} effectively captures and measures the relative similarity between different pairs.

\subsubsection{Concurrent prediction on MCS and GED similarities} This experiment evaluates the effectiveness of \textsc{G2R} in approximating GED similarity using disjoint parts, and assesses the performance of \textsc{G2R-Dual} in simultaneously predicting MCS and GED similarities. Table \ref{tab:results_mcs} and Table \ref{tab:results_ged} summarize the results for score prediction and ranking for MCS and GED similarities, respectively. The results demonstrate that both \textsc{G2R} and \textsc{G2R-Dual} consistently outperform the baselines in computing MCS and GED similarities. The decoupling of inputs for prediction proves effective, with no clear winner between \textsc{G2R} and \textsc{G2R-Dual}, except for the \emph{AIDS700} dataset in MCS similarity learning. This discrepancy is likely due to the presence of node labels in GED similarity computations, which introduces perturbations in label-less MCS similarity learning. Furthermore, we observe that training GREED on GED similarity is not affected by initialization, in contrast to its training on MCS similarity in Table \ref{tab:mcs}. This suggests that the inductive bias employed by GREED is unsuitable for MCS similarity. A similar trend is evident when training the LMCCS and XMCS models on GED similarity, since they were originally designed for MCS.

\subsubsection{Ablation Study}
To quantify the importance of different components, we propose four \textsc{G2R} variants and two \textsc{G2R-Dual} variants: (1) \textsc{w/o PE}: we disable the relative position; (2) \textsc{w/o Vol Score}: we predict similarity without considering the volume score; (3) \textsc{w/o Shape Score}: we predict similarity without considering the shape score; (4) \textsc{w/o Clamp}: we predicte similarity without clamping the graph region; (5) \textsc{w/o Uncertainty}: we train the \textsc{G2R-Dual} model using a basic dual loss; (6) \textsc{Shared MLP}: we replace $\operatorname{MLP}_{MCS}(\cdot)$ and $\operatorname{MLP}_{GED}(\cdot)$ with a shared MLP and predict the shape score for GED similarity using $\widehat{\mathbf{Score}}_s(\tilde{\mathbf{R}}_{G_1}, \tilde{\mathbf{R}}_{G_2}) = \lambda \cdot \operatorname{MLP}(\cdot)$ for the \textsc{G2R-Dual} model.

Results in Table \ref{tab:results_ged} and Table \ref{tab:results_mcs} show that incorporating relative position leads to a reduction in MSE of up to $58\%$ for the \emph{AIDS700} and \emph{Linux} datasets. However, this improvement is negligible for the \emph{IMDB-Multi} dataset, likely due to the limited graph diameter (maximum 2), where all nodes exhibit similar flow paths. This results in a reduced impact and over-smoothing effect of relative positions. This observation also explains the performance drop in MCS similarity ranking on the \emph{IMDB-Binary} dataset. Another indispensable component of \textsc{G2R} is the clamp operation, which contributes to a decrease in MSE of up to $62\%$.  However, its impact on the \emph{IMDB-Multi} dataset is limited for the same reason. Both the shape and volume scores significantly contribute to the overall performance, with the shape score generally providing greater benefits. However, for MCS similarity learning on the \emph{AIDS700} dataset, the volume score proves to be more advantageous. The uncertainty-weighted loss slightly improves \textsc{G2R-Dual}'s performance. However, even without this loss (\textsc{w/o Uncertainty}), \textsc{G2R-Dual} still outperforms the baselines and achieves performance comparable to when the loss function is used. This suggests that the simultaneous prediction of MCS and GED, achieved by decoupling the input for prediction, does not rely heavily on the optimization of a multi-task learning loss. Moreover, using a shared MLP in \textsc{G2R-Dual} does not result in a noticeable decline in performance, indicating potential for cost reduction.

\begin{figure}[t]
\begin{subfigure}{0.49\linewidth}
\includegraphics[width=\linewidth]{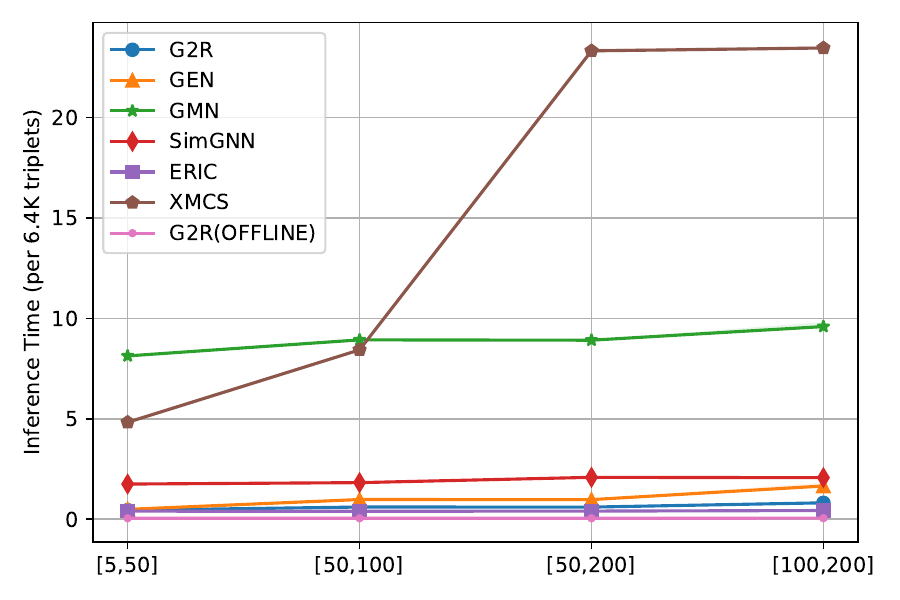}
\caption{\footnotesize Inference phase}\label{fig:inference}
\end{subfigure}
\begin{subfigure}{0.49\linewidth}
\includegraphics[width=\linewidth]{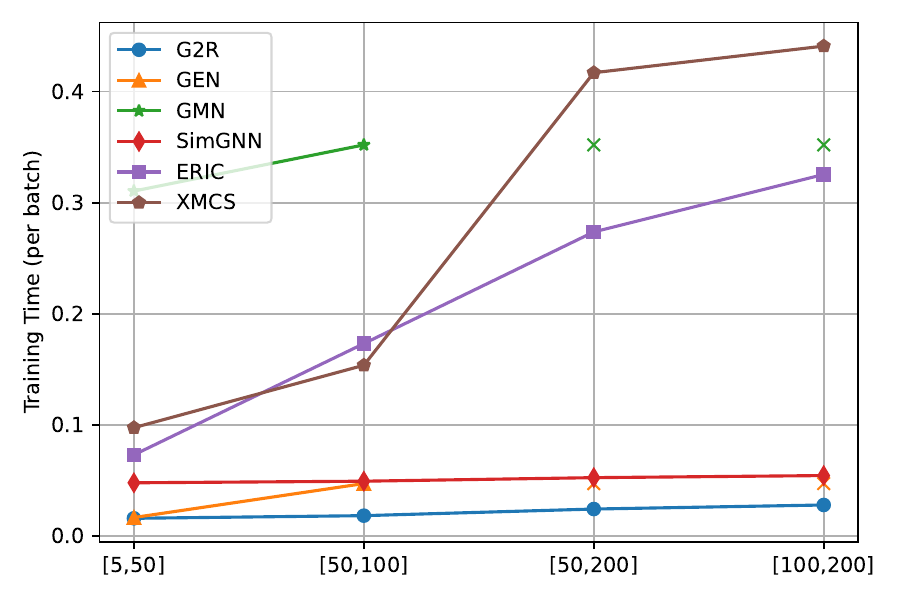}
\caption{\footnotesize Training phase}\label{fig:training}
\end{subfigure}
\caption{Time efficiency in training and inference phases (sec). X indicates out-of-memory.}\label{fig:train_inf}
\end{figure}

\begin{figure}[t]
\begin{subfigure}{0.49\linewidth}
\includegraphics[width=\linewidth]{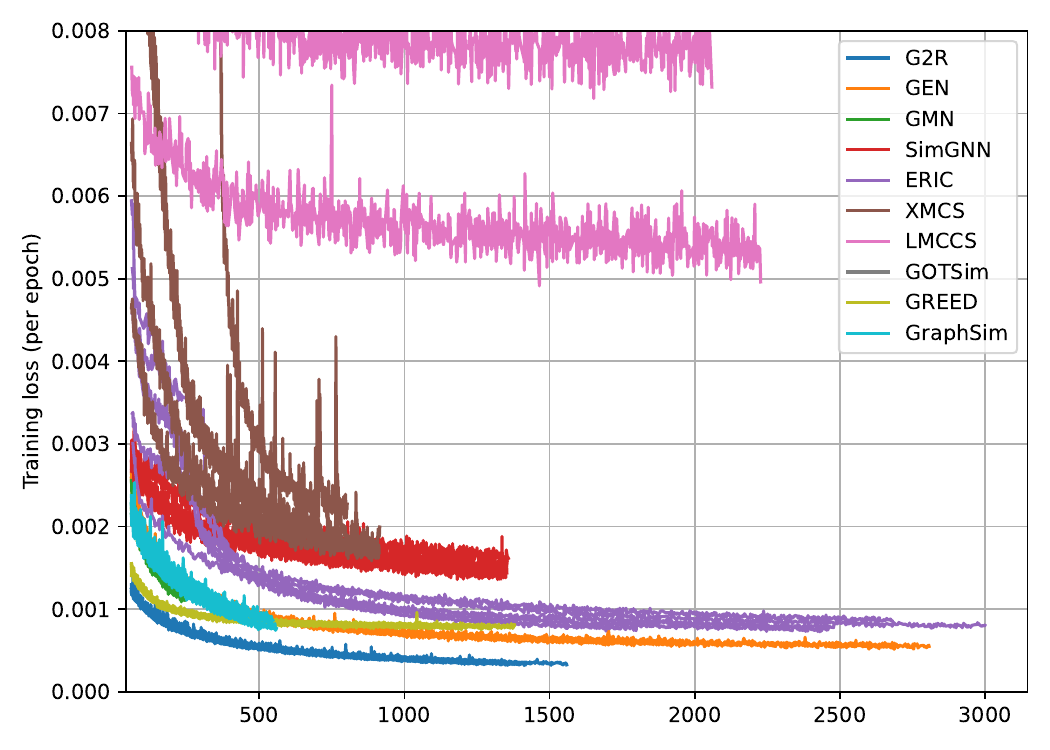}
\caption{\footnotesize Training Loss}
\end{subfigure}
\begin{subfigure}{0.49\linewidth}
\includegraphics[width=\linewidth]{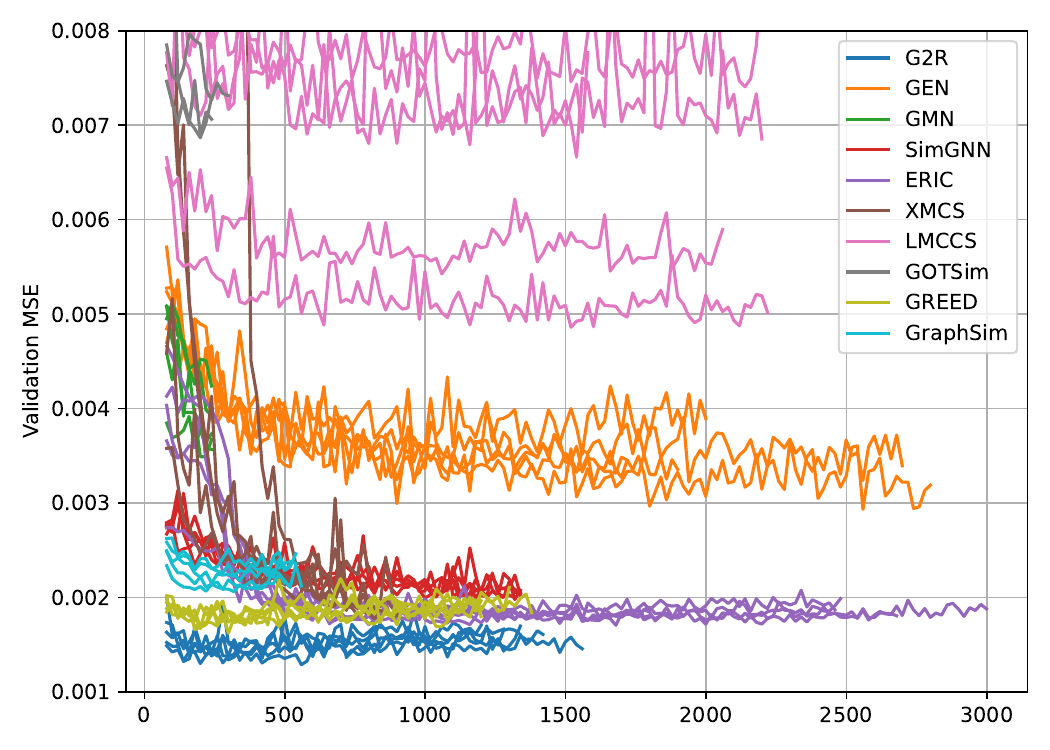}
\caption{\footnotesize Validation MSE}
\end{subfigure}
\caption{Convergence speed on GED similarity}\label{fig:converage_ged}
\end{figure}

\begin{figure}[t]
\begin{subfigure}{0.49\linewidth}
\includegraphics[width=\linewidth]{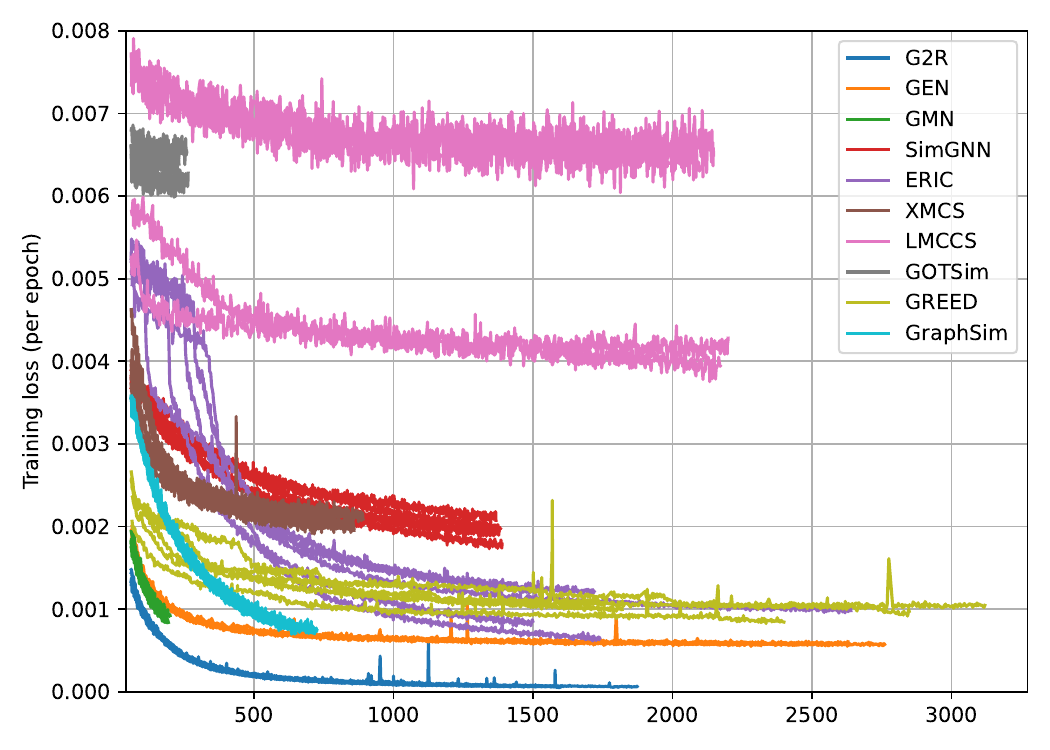}
\caption{\footnotesize Training Loss}
\end{subfigure}
\begin{subfigure}{0.49\linewidth}
\includegraphics[width=\linewidth]{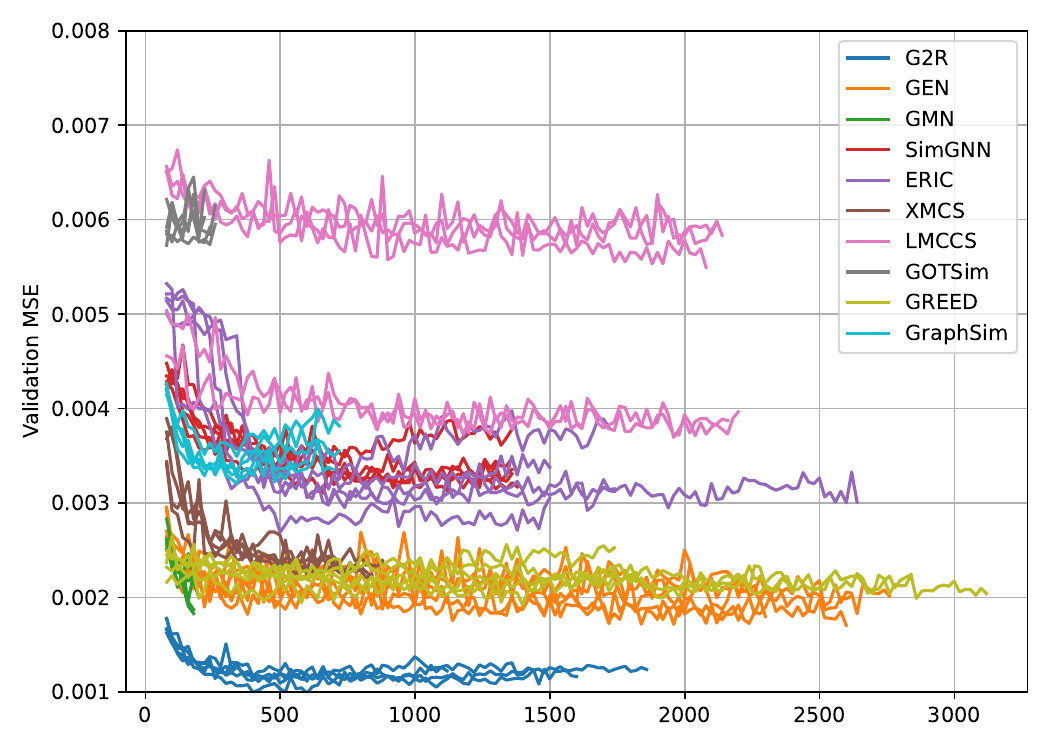}
\caption{\footnotesize Validation MSE}
\end{subfigure}
\caption{Convergence speed on MCS similarity}\label{fig:converage_mcs}
\end{figure}

\subsubsection{Time Efficiency}
We evaluate the time efficiency of the models across various node size groups, regarding: \textbf{(1) Training and Inference Time.} Results in Fig. \ref{fig:train_inf} show that, despite a slight increase in training and inference time, \textsc{G2R} remains efficient in both phases. While it may be slower than Eric during inference, this is due to \textsc{G2R}'s use of the Multi-sink Propagation mechanism. Since \textsc{G2R}'s graph representations are pair-independent, they can be stored for offline inference, where \textsc{G2R} exhibits comparable complexity to offline Eric. In contrast, the fine-grained comparison strategy significantly slows down ERIC, XMCS, and GMN during the training phase. SimGNN, however, is less affected due to its simpler architecture. It produces non-differentiable pairwise node similarity histograms, which do not require backpropagation or a refinement process. On the other hand, XMCS suffers from the iterative refinement mechanism of the Gumbel-Sinkhorn network, which drastically increases inference time and limits its scalability. \textbf{(2) Convergence Speed.} In our experiment, we limit the training time to an eight-hour time frame and analyze the convergence speed of \textsc{G2R} and other models. The results in Fig. \ref{fig:converage_ged} and \ref{fig:converage_mcs} show that \textsc{G2R} converges the fastest and achieves the lowest training loss after the warm-up phase in both GED and MCS similarity computations. Another model with good convergence speed is GREED in GED similarity learning. While GEN achieves a lower training loss with more epochs, it suffers the most from overfitting.

\begin{figure}[t]
\begin{subfigure}{\linewidth}
\includegraphics[width=\linewidth]{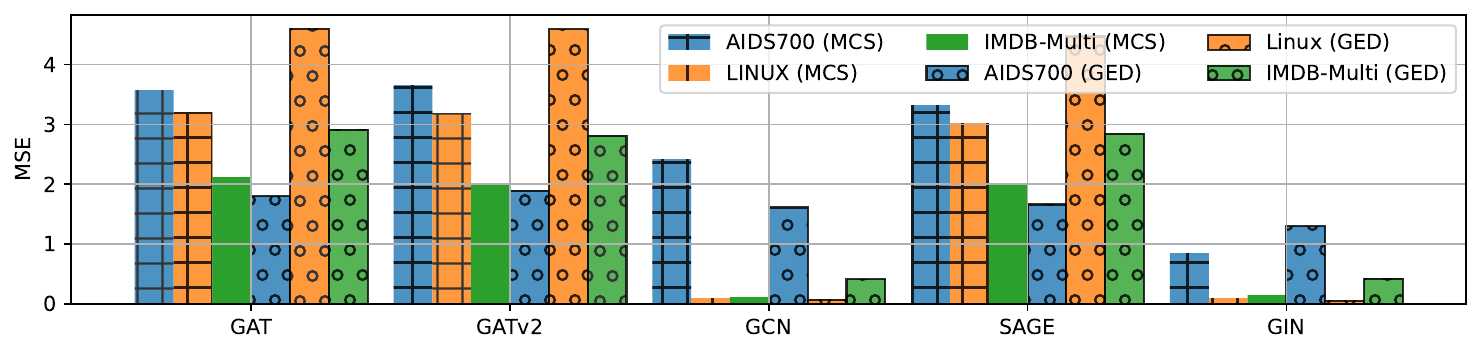}
\caption{\footnotesize GNN Backbones}
\end{subfigure}

\begin{subfigure}{0.49\linewidth}
\includegraphics[width=\linewidth]{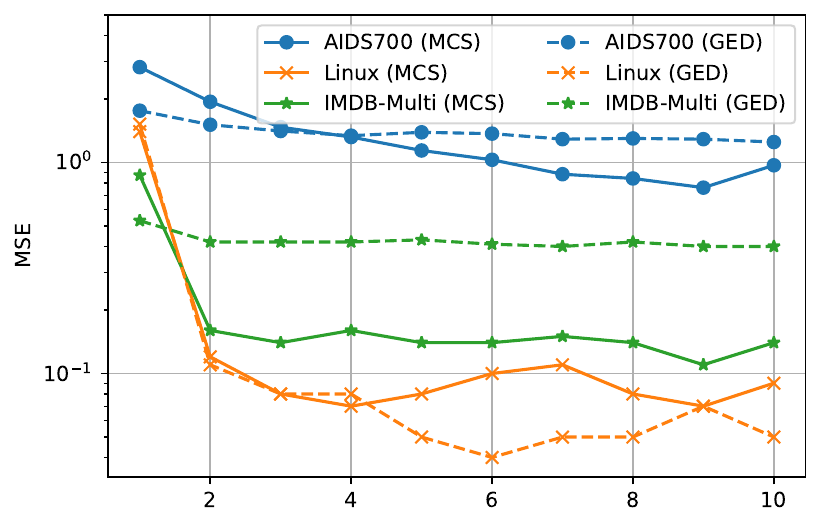}
\caption{\footnotesize Number of Layers}
\end{subfigure}
\begin{subfigure}{0.49\linewidth}
\includegraphics[width=\linewidth]{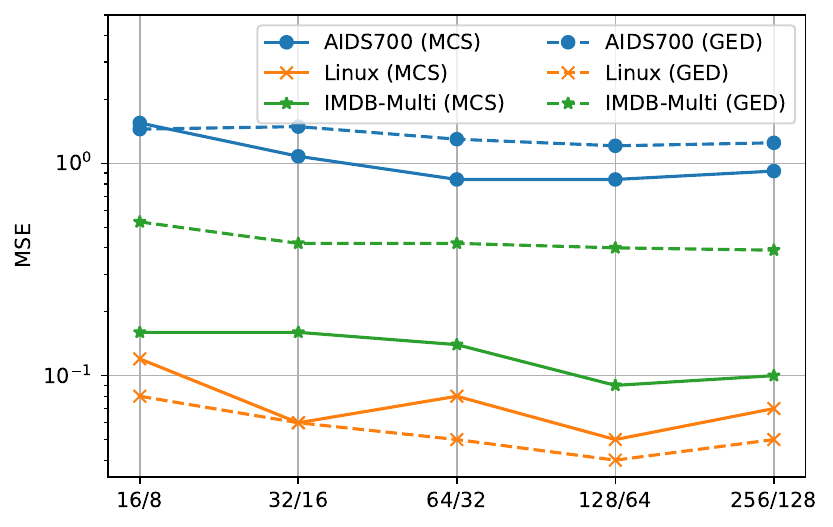}
\caption{\footnotesize Number of Hidden/Output}
\end{subfigure}

\begin{subfigure}{0.49\linewidth}
\includegraphics[width=\linewidth]{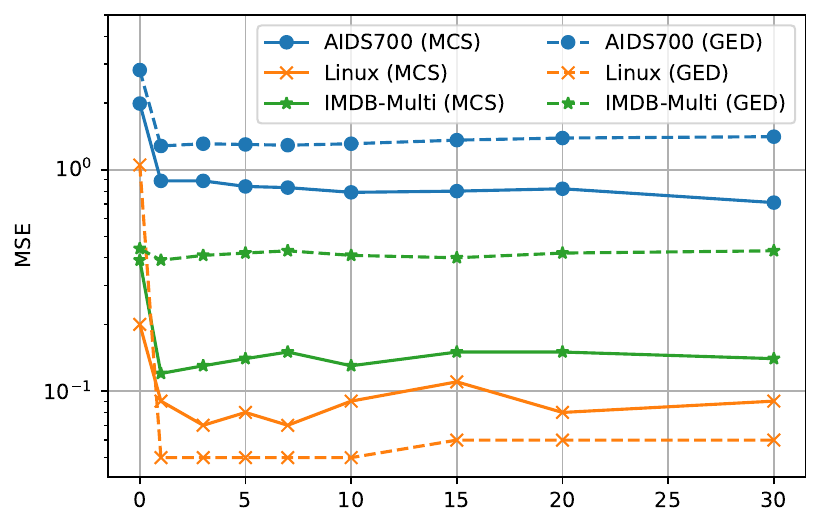}
\caption{\footnotesize Number of Flow Paths}
\end{subfigure}
\begin{subfigure}{0.49\linewidth}
\includegraphics[width=\linewidth]{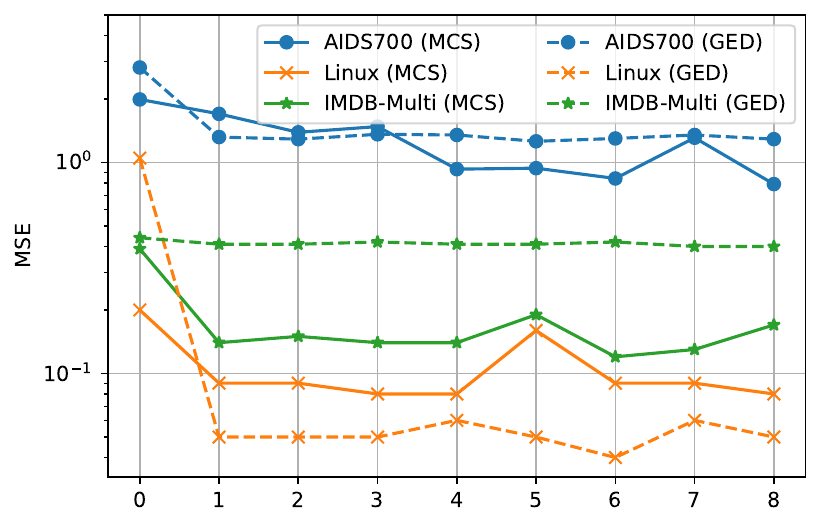}
\caption{\footnotesize Length of Flow Paths}
\end{subfigure}
\caption{Hyperparameter Sensitivity}\label{fig:hyper}
\end{figure}

\subsubsection{Hyperparameter Sensitivity} We analyze the hyperparameter sensitivity of our proposed \textsc{G2R}. For the Node-to-Region encoder, we focus on the GNN backbones, number of layers, and dimensionality. For the Multi-sink Propagation mechanism, we focus on the number and length of flow paths. Results in Fig. \ref{fig:hyper} show that, for encoder, GIN is the best backbone for \textsc{G2R}, with GCN performing second. Notably, most GED/MCS similarity computation models also use these two GNNs as backbones, indicating that these GNNs' inductive biases are well-suited for this task. \textsc{G2R} performance generally improves with more GNN layers, achieving satisfactory results with 2-4 layers and peak performance with 6-8 layers. Additionally, \textsc{G2R}'s performance increases with dimensionality, peaking at 128/64. For the Multi-sink Propagation mechanism, a small number and short length of flow paths can already significantly contribute to \textsc{G2R}'s performance. \textsc{G2R}, with 1 flow path of length 4 for each node, can deliver satisfactory results. Further performance fluctuations may depend on the specific graph properties. We hypothesize that the graph diameter plays a role. On the AIDS700 dataset, which has the largest graph diameter, \textsc{G2R} showed improved performance as the length and number of flow paths increased.

\subsubsection{Interpretability}

\begin{table}[]
\caption{We have examined $\alpha_1$ and $\beta_1$, the weights assigned to shape and volume score for MCS similarity, $\alpha_2$ and $\beta_2$, the weights assigned to shape and volume score for GED similarity, and $\gamma$, the weight that imitates the positive correlation between Bunke GED and GED.}\label{tab:weights}
\resizebox{\linewidth}{!}{
\begin{tabular}{lccccc}
\toprule
\textbf{Dataset} &
$\alpha_1$ &
$\beta_1$ &
$\alpha_2$ &
$\beta_2$ &
$\gamma$ \\
\midrule
\textsc{AIDS700} &
1.20 $\pm$ 0.01 &
0.96 $\pm$ 0.05 & 
1.11 $\pm$ 0.06 &  
0.89 $\pm$ 0.04 &   
1.83 $\pm$ 0.05 \\
\textsc{Linux} & 
1.15 $\pm$ 0.02 &                 
0.97 $\pm$ 0.06 &                 
0.97 $\pm$ 0.09 &                 
1.01 $\pm$ 0.01 &                   
1.22 $\pm$ 0.07 \\
\textsc{IMDB-Multi} &
1.06 $\pm$ 0.05 &                   
0.97 $\pm$ 0.06 &                   
0.71 $\pm$ 0.04 &                     
1.01 $\pm$ 0.00 &
4.72 $\pm$ 0.14 \\
\bottomrule
\end{tabular}}
\end{table}

We evaluate the importance of shape and volume scores, along with the learned $\gamma$, in approximating GED similarity. Table \ref{tab:weights} shows that shape scores receive higher weights than volume scores, aligning with our ablation study, which found shape scores provide a greater performance boost. The positive $\gamma$ across datasets further supports the correlation between Bunke GED and GED, confirming the positive relationship between Bunke GED and its upper bound.

\begin{figure}[t]
\begin{minipage}{1\linewidth}
    \begin{subfigure}{1\linewidth}
        \includegraphics[width=\linewidth]{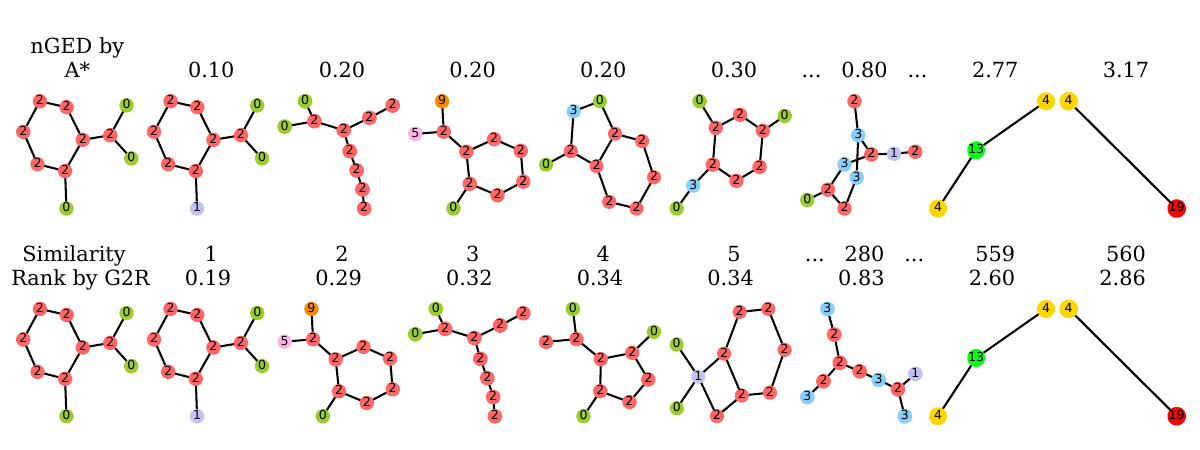}
        \caption{\footnotesize GED Ranking Results}
    \end{subfigure}\\
    \begin{minipage}{1\linewidth}
        \begin{subfigure}{0.19\linewidth}
            \includegraphics[width=\linewidth]{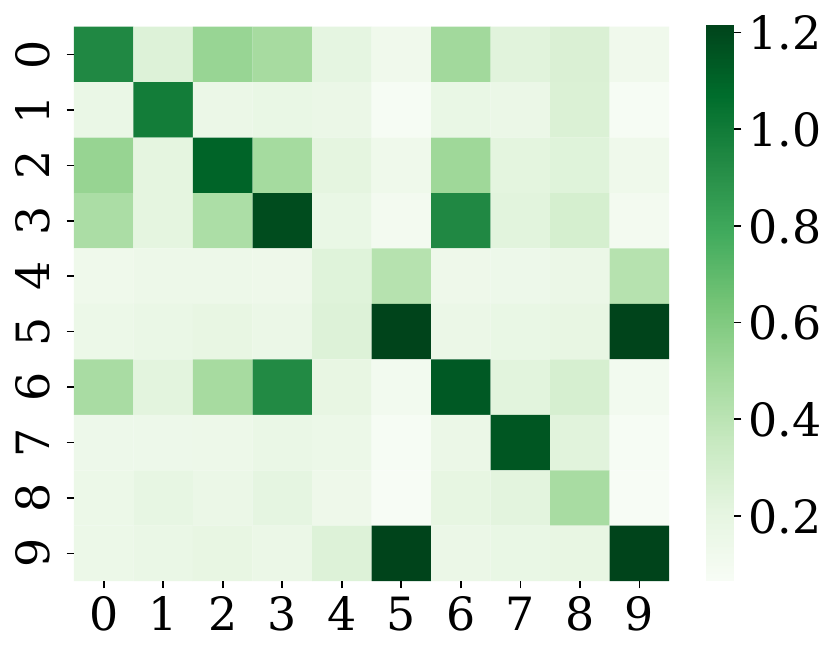}
            \caption{\footnotesize $1^{\text{st}}$}
        \end{subfigure}
        \begin{subfigure}{0.19\linewidth}
            \includegraphics[width=\linewidth]{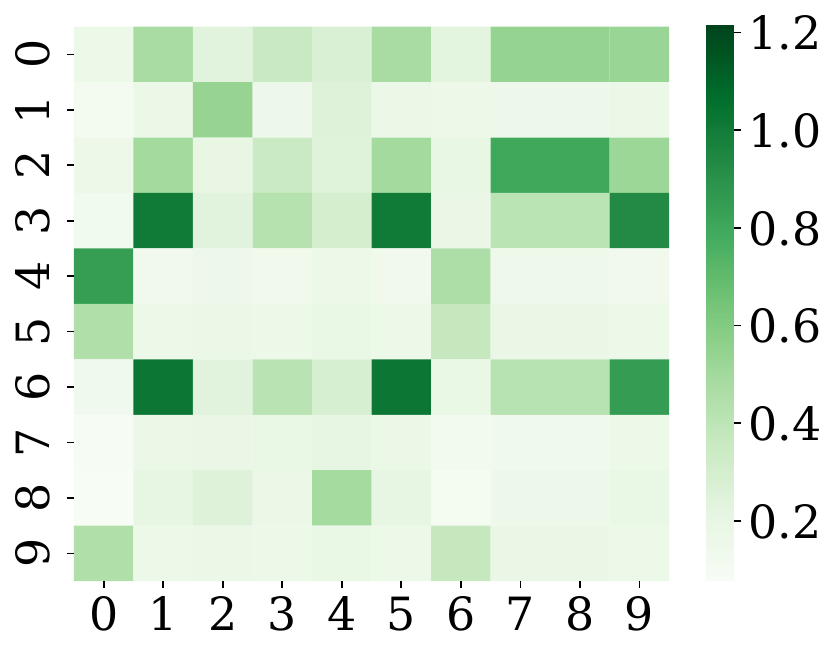}
            \caption{\footnotesize $10^{\text{th}}$}
        \end{subfigure}
        \begin{subfigure}{0.19\linewidth}
            \includegraphics[width=\linewidth]{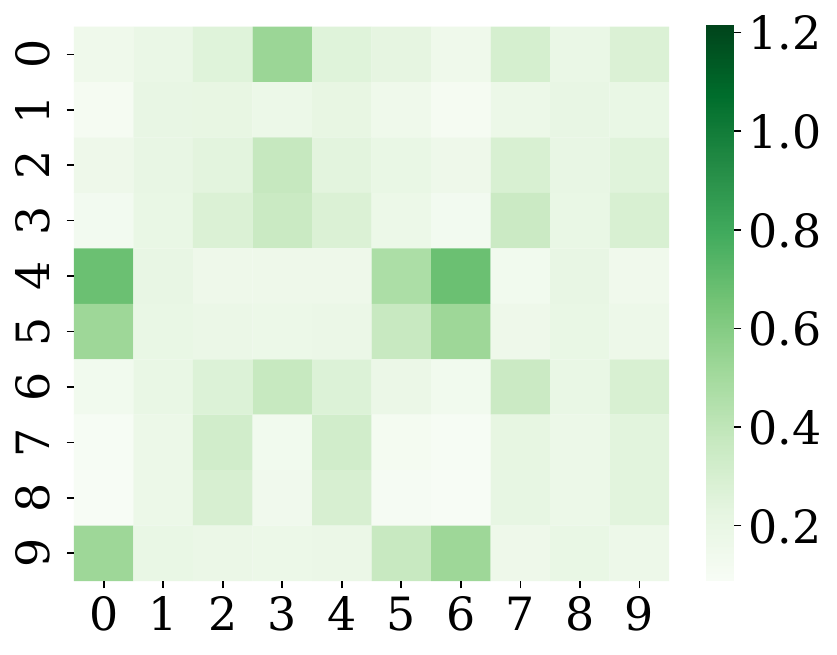}
            \caption{\footnotesize $50^{\text{th}}$}
        \end{subfigure}
        \begin{subfigure}{0.19\linewidth}
            \includegraphics[width=\linewidth]{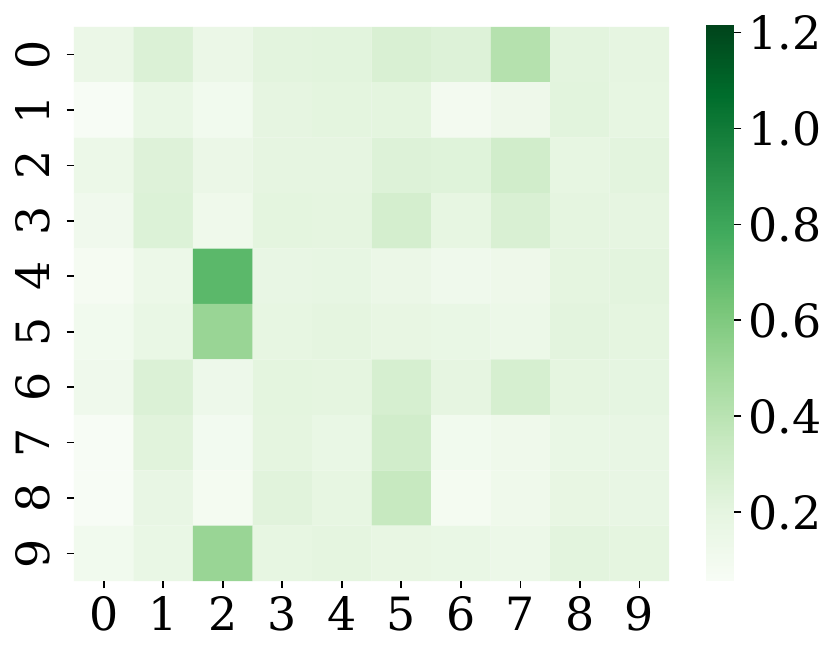}
            \caption{\footnotesize $280^{\text{th}}$}
        \end{subfigure}
        \begin{subfigure}{0.19\linewidth}
            \includegraphics[width=\linewidth]{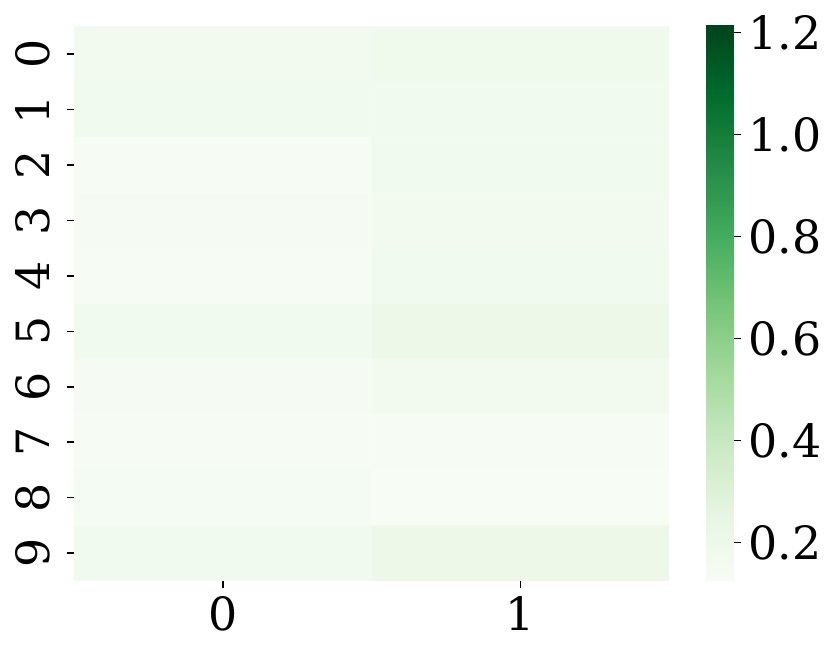}
            \caption{\footnotesize the last}
        \end{subfigure}
        \caption{Heatmap of pairwise node similarity for GED}\label{fig:ged}
    \end{minipage}
\end{minipage}\\
\begin{minipage}{1\linewidth}
\begin{subfigure}{1\linewidth}
\includegraphics[width=\linewidth]{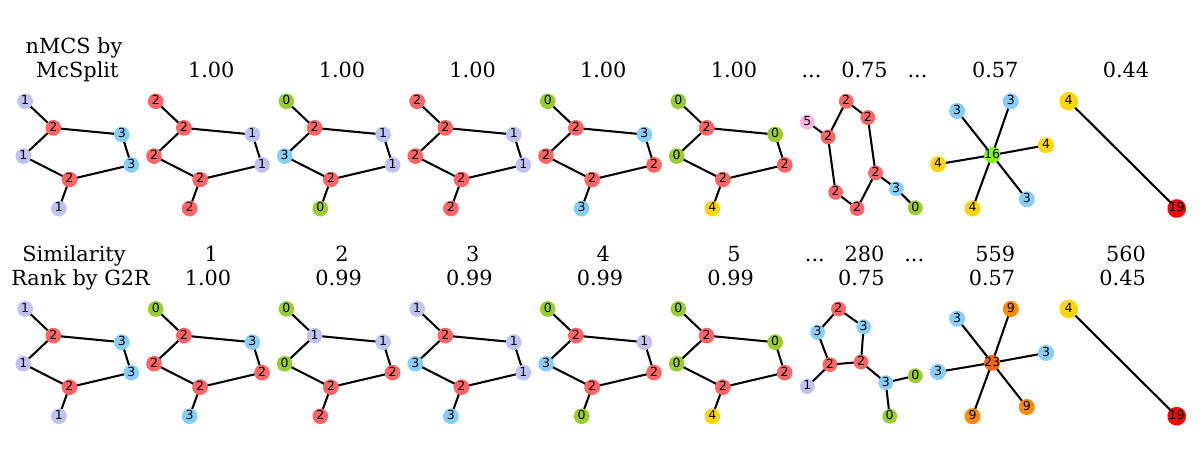}
\caption{\footnotesize MCS Ranking Results}
\end{subfigure}
\begin{minipage}{1\linewidth}
        \begin{subfigure}{0.19\linewidth}
            \includegraphics[width=\linewidth]{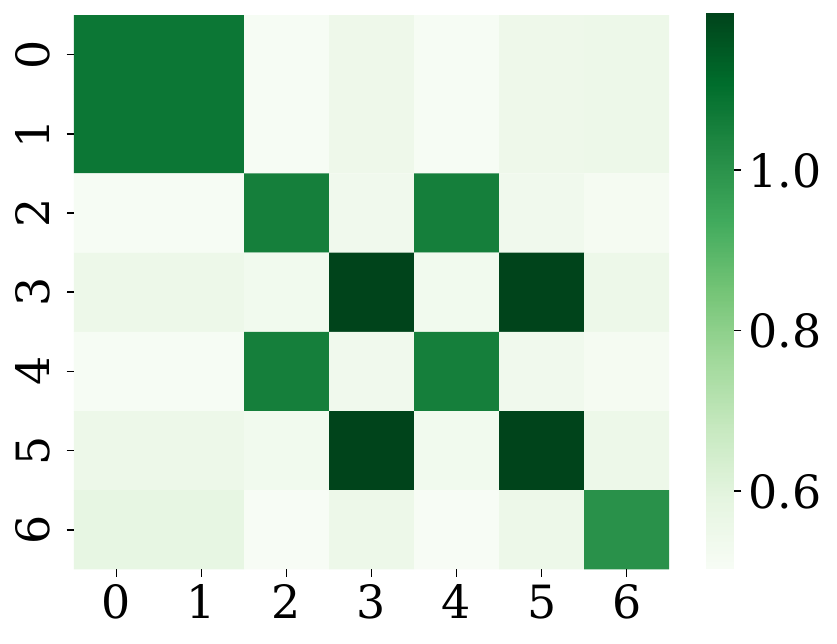}
            \caption{\footnotesize $1^{\text{st}}$}
        \end{subfigure}
        \begin{subfigure}{0.19\linewidth}
            \includegraphics[width=\linewidth]{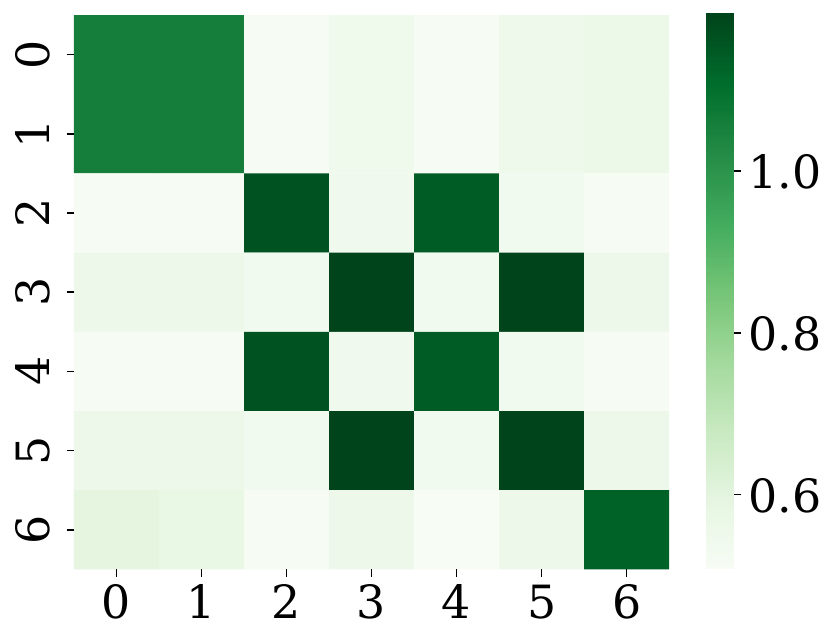}
            \caption{\footnotesize $10^{\text{th}}$}
        \end{subfigure}
        \begin{subfigure}{0.19\linewidth}
            \includegraphics[width=\linewidth]{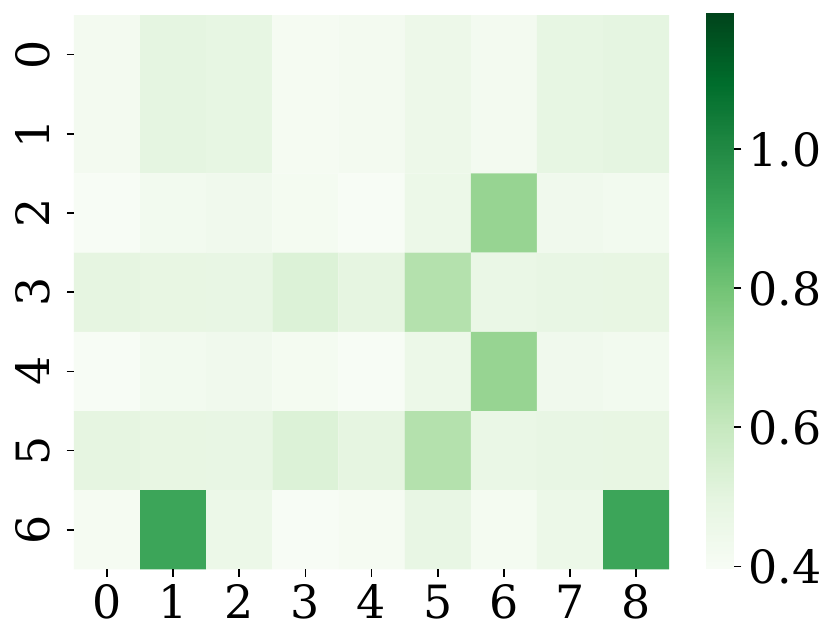}
            \caption{\footnotesize $50^{\text{th}}$}
        \end{subfigure}
        \begin{subfigure}{0.19\linewidth}
            \includegraphics[width=\linewidth]{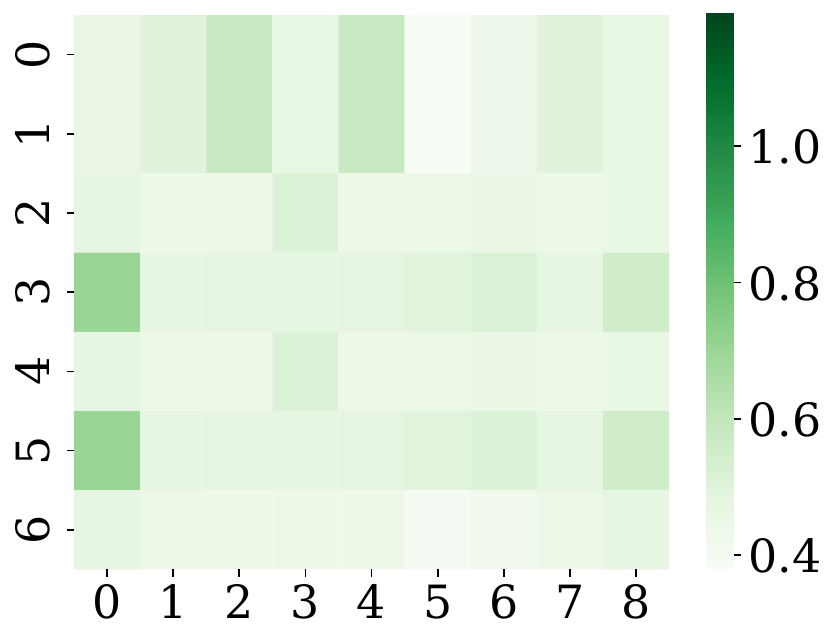}
            \caption{\footnotesize $280^{\text{th}}$}
        \end{subfigure}
        \begin{subfigure}{0.19\linewidth}
            \includegraphics[width=\linewidth]{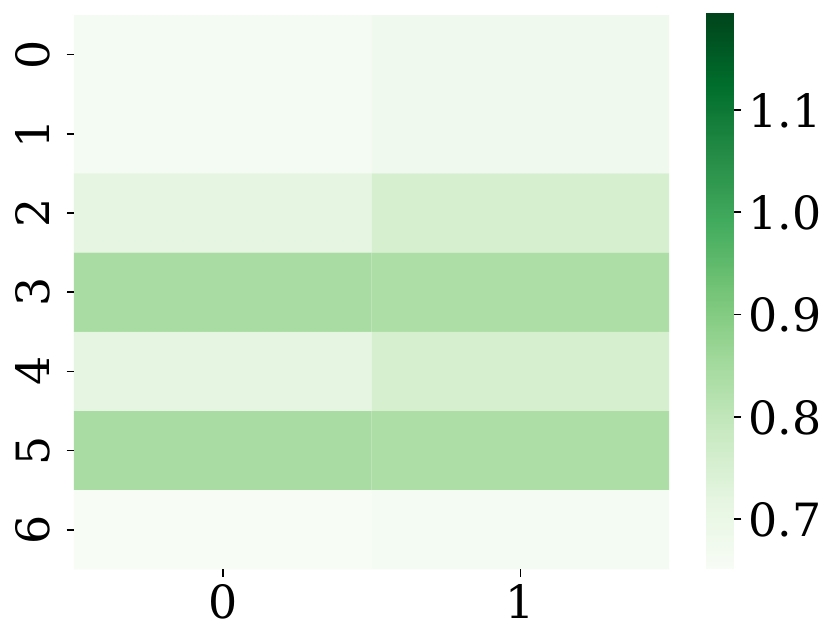}
            \caption{\footnotesize the last}
        \end{subfigure}
        \caption{Heatmap of pairwise node similarity for MCS}\label{fig:mcs}
    \end{minipage}
\end{minipage}
\end{figure}

\subsubsection{Case Study} Figures \ref{fig:ged} and \ref{fig:mcs} show the GED and MCS rankings generated by $\textsc{G2R}$ for two queries, with ground-truth rankings calculated using exact algorithms. The ranking results demonstrate strong consistency between $\textsc{G2R}$ and the ground truth, indicating high ranking accuracy. Additionally, the heatmap of pairwise node similarity, computed based on shape and volume of overlap regions, shows strong correlations between matched nodes, suggesting that $\textsc{G2R}$'s intermediate representations can effectively guide node alignment. 

\section{Conclusion}
We present \textsc{Graph2Region} (\textsc{G2R}), a novel graph embedding method for efficient graph similarity learning. \textsc{G2R}  leverages closed regions to represent nodes and captures adjacency patterns through a Multi-sink Propagation mechanism. \textsc{G2R} surpasses existing approaches by explicitly restoring structural and scale information to enhance the expressiveness of graph embeddings. Moreover, \textsc{G2R} predicts MCS similarity based on the overlap of graph regions and uses disjoint regions as a proxy for GED similarity, which empowers \textsc{G2R} with the unique capability of concurrently predicting MCS and GED similarity. Extensive experiments on 15 datasets validate the effectiveness, transferability, and time efficiency of \textsc{G2R}.

\begin{IEEEbiographynophoto}
{Zhouyang Liu} received the BSc (2013-2016) and MSc (2016-2018) degrees in computer science from the Université of Franche-Comté in Besançon, France. She is currently pursuing a PhD in the College of Computer Science and
Technology at the National University of Defense Technology in Changsha, Hunan, China. Her research focuses on graph similarity computation, and graph representation learning.
\end{IEEEbiographynophoto}
\begin{IEEEbiographynophoto}
{Yixin Chen} received the B.E. degree in computer science from the National University of Defense Technology in 2009, and the Ph.D. degree from the School of Computer Science at McGill University in 2018. He is currently a research associate professor at the National University of Defense Technology. His research interests include graph neural networks and multimedia big data systems.
\end{IEEEbiographynophoto}
\begin{IEEEbiographynophoto} 
{Ning Liu} received the PhD degree in computer science from the College of Computer Science and Technology, National University of Defense Technology, Changsha, China, in 2023. She is currently an assistant research fellow with the Information Support Force Engineering University. Her research interests include graph representation learning, graph OOD generalization, and graph anomaly detection.
\end{IEEEbiographynophoto}
\begin{IEEEbiographynophoto}
{Jiezhong He} is currently pursuing a Ph.D. in Computer Science and Technology at the College of Computer, National University of Defense Technology, Changsha, China. His research interests include subgraph matching and graph processing systems.
\end{IEEEbiographynophoto}
\begin{IEEEbiographynophoto}
{Dongsheng Li}
received the BSc (with honors) and PhD (with honors) degrees in computer science from the College of Computer, National University
of Defense Technology, Changsha, China, in 1999 and 2005, respectively. He was awarded the prize of National Excellent Doctoral Dissertation of PR China by the Ministry of Education of China in 2008. He is now a full professor at the National Lab for Parallel and Distributed Processing, National University of Defense Technology, China. His research interests include parallel and distributed computing, cloud computing, and large-scale data management.
\end{IEEEbiographynophoto}

\end{document}